\documentclass[10pt,twocolumn,letterpaper]{article}

\usepackage{cvpr}
\usepackage{times}
\usepackage{epsfig}
\usepackage{graphicx}
\usepackage{amsmath}
\usepackage{amssymb}
\usepackage{sidecap}
\usepackage{array}
\usepackage{caption}
\usepackage{subfigure}
\usepackage{url}
\usepackage{enumitem}

\usepackage[breaklinks=true,bookmarks=false]{hyperref}

\cvprfinalcopy 

\ifcvprfinal\fi

\begin{document}

\title{Pose-Aware Person Recognition}

\author{Vijay Kumar $^ \star$  \qquad Anoop Namboodiri $^ \star$ \qquad Manohar Paluri $^ \dagger$ \qquad  C. V. Jawahar $^ \star$\\
\qquad $^\star$ CVIT, IIIT Hyderabad, India \qquad $^\dagger$ Facebook AI Research\\
}

\maketitle
\thispagestyle{empty}



\begin{abstract}
   Person recognition methods that use multiple body regions have shown significant improvements over traditional face-based recognition. One of the primary challenges in full-body person recognition is the extreme variation in pose and view point. In this work, (i) we present an approach that tackles pose variations utilizing multiple models that are trained on specific poses, and combined using pose-aware weights during testing. (ii) For learning a person representation, we propose a network that jointly optimizes a single loss over multiple body regions. (iii) Finally, we introduce new benchmarks to evaluate person recognition in diverse scenarios and show significant improvements over previously proposed approaches on all the benchmarks including the photo album setting of PIPA.
\end{abstract}
\section{Introduction}
 People are ubiquitous in our images and videos. They appear in photographs, entertainment videos, sport broadcasts, and surveillance and authentication systems. This makes person recognition an important step towards automatic understanding of media content. Perhaps, the most straight-forward and popular way of recognizing people is through facial cues. Consequently, there is a large literature focused on face recognition~\cite{lfw_survey, Face_Recognition_Literature_Survey}. Current face recognition algorithms~\cite{deepface_omkar, facenet, deepID03, deepface} achieve impressive performance on verification and recognition benchmarks~\cite{lfw,youtube_faces}, and are close to human-level performance.

\begin{figure}
\centering
\includegraphics[scale=0.5]{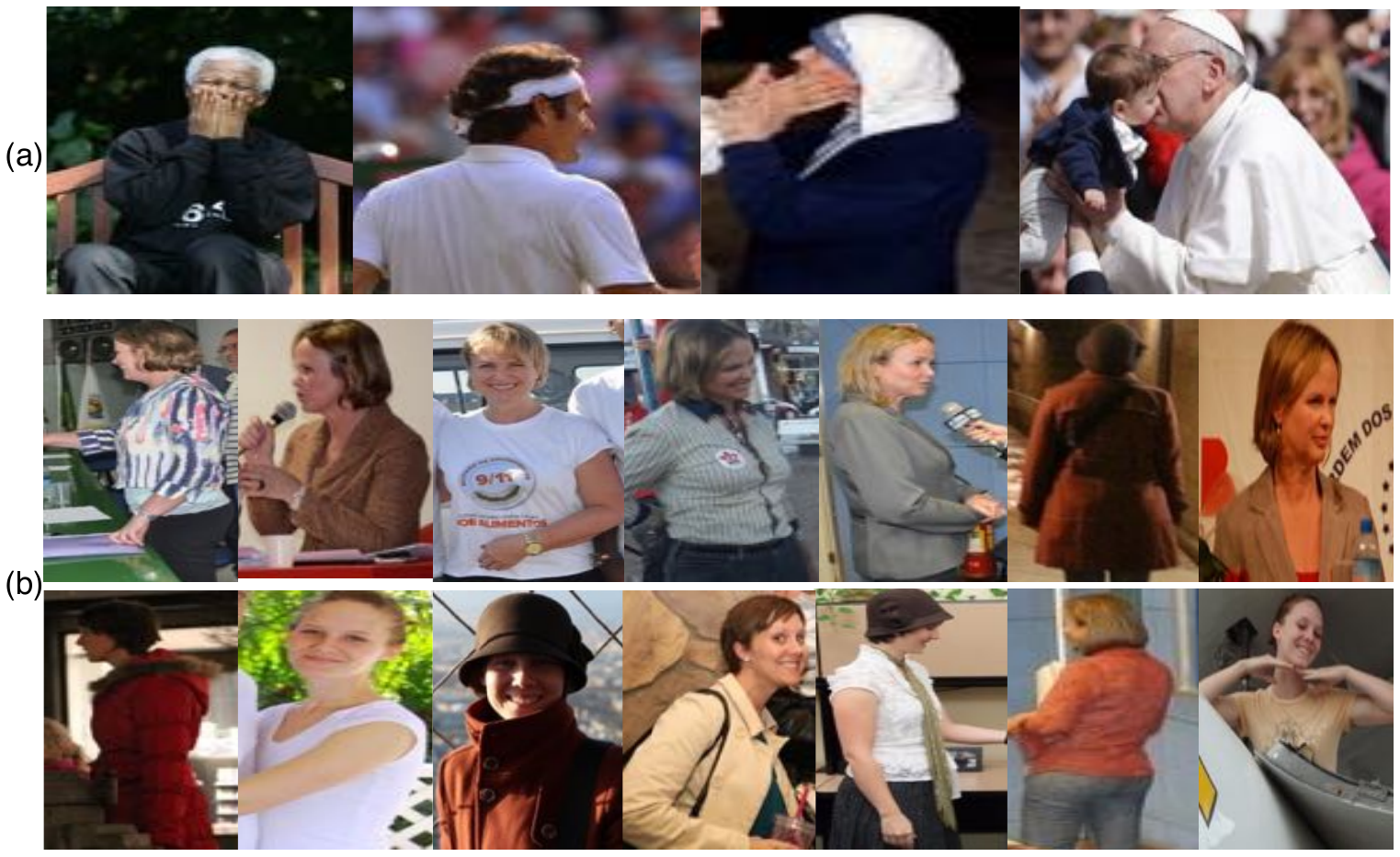}
\vspace*{-1mm}
\caption{(a) Different body regions provide cues that help in recognition. (b) Each row shows the appearance of a person in different poses. Note that the distinguishing features of a person that help classification are different in each pose.}
\label{fig:pose_intro}
\end{figure}

Face based recognition approaches require that faces are visible in images, which is often not the case in many practical scenarios. For instance, in social media photos, movies and sport videos, faces may be occluded, be of low resolution, facing away from the camera or even cropped from the view (Figure~\ref{fig:pose_intro}(a)). Hence it becomes necessary to look {\em beyond faces} for additional identity cues. It has been recently demonstrated~\cite{person_recognition_bodyparts, person_recognition_context, person_recognition_poselets} that different body parts provide complementary information and significantly improve person recognition when used in conjunction with face.

One of the major challenges in person recognition or fine-grained object recognition in general is the pose and alignment of different object parts. The appearance of the same object changes drastically with different poses and view-points (rows of Figure~\ref{fig:pose_intro}(b)) causing a serious challenge for recognition. One way to overcome this problem is through pose normalization, where objects in different poses and view-points are transformed to a canonical pose \cite{ face_alignment01,face_alignment05,face_alignment02,deepface,face_alignment03,face_alignment06}. Another popular strategy is to model the appearance of objects in individual poses by learning view-specific representations \cite{pose_aware_face02, pose_aware_face03, pose_aware_face01, panda}.

In this work, we aim to learn pose-aware representations for person recognition. While it is straight-forward to align objects such as faces, it is harder to align human body parts that exhibit large variations. Hence we design view-specific models to obtain pose-aware representations. We partition the space of human pose into finite clusters (columns of Figure~\ref{fig:pose_intro}(b)) each containing samples in a particular body orientation or view-point. We then learn multi-region convolutional neural network (convnet) representations for each view-point. However, unlike previous approaches that train a convnet for each body region, we jointly optimize the network over multiple body regions with a single identification loss. This provides additional flexibility to the network to make predictions based on a few informative body regions. This is in contrast to separate training which strictly enforces correct predictions from each body region. 
During testing, we obtain the identity predictions of a sample through a linear combination of classifier scores, each of which is trained using a pose-specific representation. The weights for combining the classifiers are obtained by a pose estimator that computes the likelihood of each view.

Our approach overcomes some of the limitations of the previously proposed approaches, {\tt PIPER}~\cite{person_recognition_poselets} and {\tt naeil}~\cite{person_recognition_bodyparts}. Although poselet-based representation of {\tt PIPER} normalizes the pose; individual poselet patches \cite{poselets} by themselves are not discriminative enough for recognition tasks, and under-perform compared to fixed body regions such as head and upper body. On the other hand, {\tt naeil} learns a pose-agnostic representation using more informative body regions. Our framework is able to combine the best of both approaches by generating pose-specifc representations based on discriminative body regions, which are combined using pose-aware weights. 

Another major contribution of the work is the rigorous evaluation of person recognition. Current approaches \cite{person_recognition_bodyparts, person_recognition_context,person_recognition_poselets} have solely focused on the photo albums scenario, reporting primarily on \textsc{PIPA} dataset \cite{person_recognition_poselets}. However, this setting is very limited due to the similar appearance of people in albums, clothing and scene cues. 
To create a more challenging evaluation, we consider three different scenarios of photo albums, movies and sports and show significant performance improvements with our proposed approach. The datasets are available at {\tt \footnotesize \url{http://cvit.iiit.ac.in/research/projects/personrecognition}}

 \section{Related Work}
 Person recognition has been attempted in multiple settings, each assuming the availability of specific types of information regarding the subjects to be recognized.

{\bf Face recognition} is by far the most widely studied form of person recognition. 
The area has witnessed great progress with several techniques proposed to solve the problem, varying from hand-crafted feature design~\cite{lbp_face, gabor_face, lbp_face2}
, metric learning~\cite{sift_face2, fisher_face}, sparse representations~\cite{SRC, dksvd} 
to state-of-the-art deep representations~\cite{deepface_omkar, facenet, deepface}. 

{\bf Person re-identification} is the task of matching pedestrians captured in non-overlapping camera views; a primary requirement in video-surveillance applications. Most popular existing works employ metric learning~\cite{person_reid00, person_reid03, person_reid01} 
using hand-crafted \cite{person_reid05, person_reid06, person_reid08} or data-driven \cite{person_reid11, person_reid09, person_reid10} features to achieve invariance with respect to view-point, pose and photometric transformations. The approaches in \cite{person_reid12,person_reid01} also optimized a joint architecture with siamese loss on non-overlapping body regions for re-identification.

{\bf Pose normalization} and {\bf multi-view representation} are the two common approaches in dealing with object pose variations.
Frontalization \cite{face_alignment02, deepface,face_alignment03,face_alignment06} is a pose normalization
scheme used commonly in face recognition, where faces in arbitrary poses are transformed to a canonical pose before recogniton. Pose-normalization is also applied to the similar problem of fine-grained bird classification~\cite{face_alignment01, bird_pose_normalization}. Unlike rigid objects such as faces, it is difficult to align human body parts due to large deformations. 
Hence we follow a multi-view representation approach where the objects are modeled independently in different views. This has also been employed in face recognition~\cite{pose_aware_face02, pose_aware_face01}, where training faces are grouped into different poses and pose-aware \textsc{CNN} representations are learnt for each group. 

{\bf Person recognition} with multiple body cues is the problem of interest in this work. We make direct comparisons with the recent efforts that use multiple body cues: {\tt PIPER}~\cite{person_recognition_poselets}, {\tt naeil}~\cite{person_recognition_bodyparts} and Li \etal~\cite{person_recognition_context}. 
{\tt PIPER} uses a complex pipeline with $109$ classifiers, each predicting identities based on different body part representations. These include one representation based on Deep Face~\cite{deepface} architecture trained on millions of images, one AlexNet~\cite{alexnet} trained on the full body and $107$ AlexNets trained on poselet patches, the latter two using \textsc{PIPA}~\cite{person_recognition_poselets} trainset. On the other hand, {\tt naeil} is based on fixed body regions such as face, head and body along with scene and human attribute cues trained using four different datasets, namely \textsc{PIPA}, \textsc{CASIA}~\cite{CASIA}, \textsc{CACD}~\cite{CACD} and \textsc{PETA}~\cite{PETA}. 
While poselets (used in {\tt PIPER}) normalizes the pose, they are less discriminative compared to fixed body regions employed by {\tt naeil}. We combine the strengths of both approaches using pose-aware representations based on fixed body regions.

\begin{figure*}
\centering
\includegraphics[width = 0.9\linewidth]{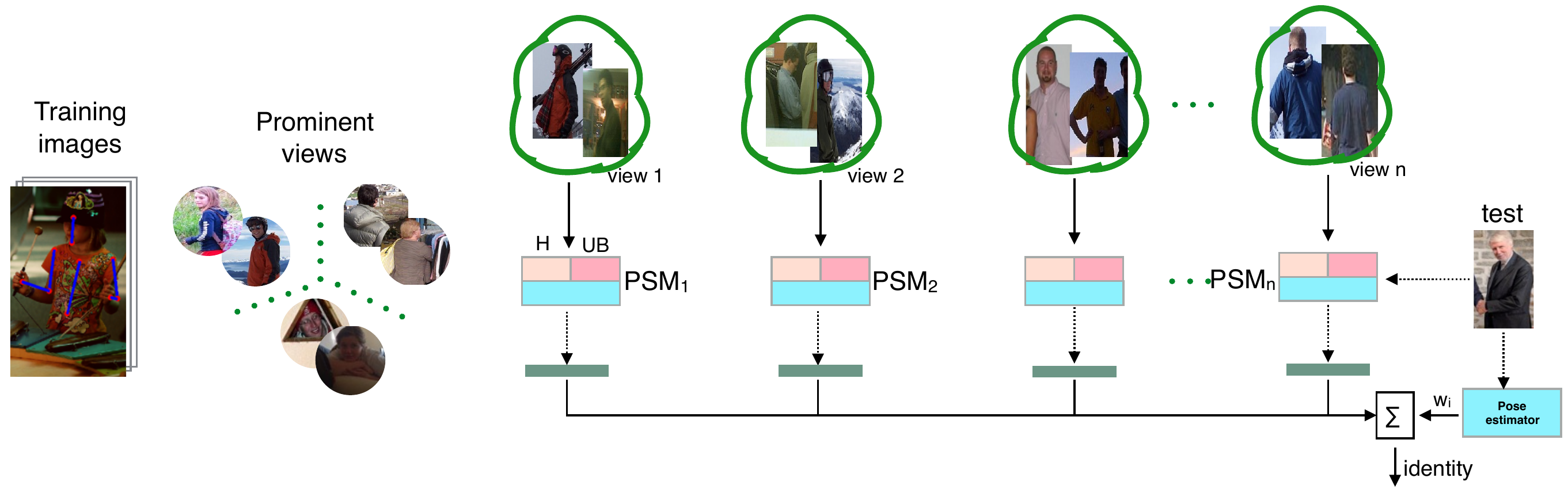}
\caption{{\bf Overview of our approach:} The database is partitioned into a set of prominent views (poses) based on keypoints. A {\it PSM} is trained for each pose based on multiple body regions. During testing, predictions from multiple classifiers, each based on a particular {\it PSM} representation, are obtained and combined using pose-aware weights provided by the pose estimator.}
\label{fig:proposed_approach}
\end{figure*}

{\bf Person identification using context} is another popular direction of work, where domain-specific information is exploited. Li \etal~\cite{person_recognition_context} focus on person recognition in photo-albums exploiting context at multiple levels.  They propose a transductive approach where a spectral embedding of the training and test examples is used to find the nearest neighbors of a test sample. An online classifier is then trained to classify each test sample. They also exploited photo meta-data such as time-stamp and the co-occurrence of people to improve the performance. 
 In \cite{face_recognition_context01, face_recognition_context02}, meta-data and clothing information are exploited to identify people in photo collections. Similarly in \cite{face_recognition_context04}, the authors use timestamp, camera-pose, and people co-occurrence to find all the instances of a specific person from a community-contributed set of photos of a crowded public event. 
 Sivic \etal~\cite{face_recognition_context03} improve the recall by modeling the appearance of cloth, hair, skin of people in repeated shots of the same scene. 

{\bf People identification in videos} may use cues such as sub-title \cite{recognition_video02} or appearance models \cite{recognition_video01, recognition_video03}, in addition to clothing, audio, face \cite{recognition_video04}. Similarly, a combination of jersey, face identification and contextual constraints are used to identify players in broadcast videos \cite{soccer01, soccer02}.

We focus on the generic person recognition problem similar to \cite{person_recognition_bodyparts, person_recognition_poselets} that work in diverse settings without using any domain level information and demonstrate the effectiveness of the pose-aware models in different scenarios. 

\section{Pose-Aware Person Recognition}
The primary challenge in person recognition is the variation in pose\footnote{In this work, we use the terms pose and view interchangeably. We use these terms to refer the overall orientation of the body with respect to the camera and not the location of keypoints within the body.} of the subjects. The appearance of the body parts change significantly with pose. We aim to tackle this by learning pose-specific models ({\it PSM}s), where each {\it PSM} focuses on specific discriminative features that are relevant to a particular pose. We fuse the information from different {\it PSM}s to make an identity prediction. 

%
Our proposed framework is shown in Figure~\ref{fig:proposed_approach}. Given a database of training images with identity labels and key-points, we cluster the images into a set of prominent views (poses) based on keypoint features (\autoref{sec:viewlearn}). A pose estimator is learned on these clusters for view classification. For learning person representation in each view, a {\it PSM} is then trained for identity recognition that makes use of multiple body regions (\autoref{sec:psmlearn}). We train multiple linear classifiers that predict the identities based on {\it PSM} representations (\autoref{sec:predict}). Given an input image $x$, we first compute the pose-specific identity scores, $s_i(y,x)$, each based on the $i^{th}$ {\it PSM} representation. The final score for each identity $y$ is a linear combination of the pose-specific scores.
	\begin{equation} 
		s(y, x) = \sum_i w_i s_i(y, x) ,
	\label{eqn1}
	\end{equation}
where $w_i$s are the pose-aware weights predicted by the pose estimator (\autoref{sec:viewlearn}). To allow robustness to rare views with limited training examples, we also incorporate a {\tt base} model in the above equation similar to \cite{person_recognition_poselets} which is trained on the entire train set, whose scores and weights are referred as $s_o(y,x)$ and $w_o$ respectively. The predicted label of the sample is computed as: $\arg\max_y s(y, x)$. 

Our framework differs from {\tt PIPER} in two aspects. First, our pose-aware weights are specific to each instance as opposed to the {\tt PIPER}, which uses fixed weights computed from a validation set. Second, {\tt PIPER} extracts features from a single model for a given localized poselet patch, however, we extract features from different pose models but combine them softly using the pose weights. This allows multiple {\it PSM}s that are very near in pose space (\eg. a semi-left and left) to contribute during the prediction.

\subsection{Learning Prominent Views}
\label{sec:viewlearn}
\begin{figure}
\centering
\includegraphics[width = 0.95\linewidth]{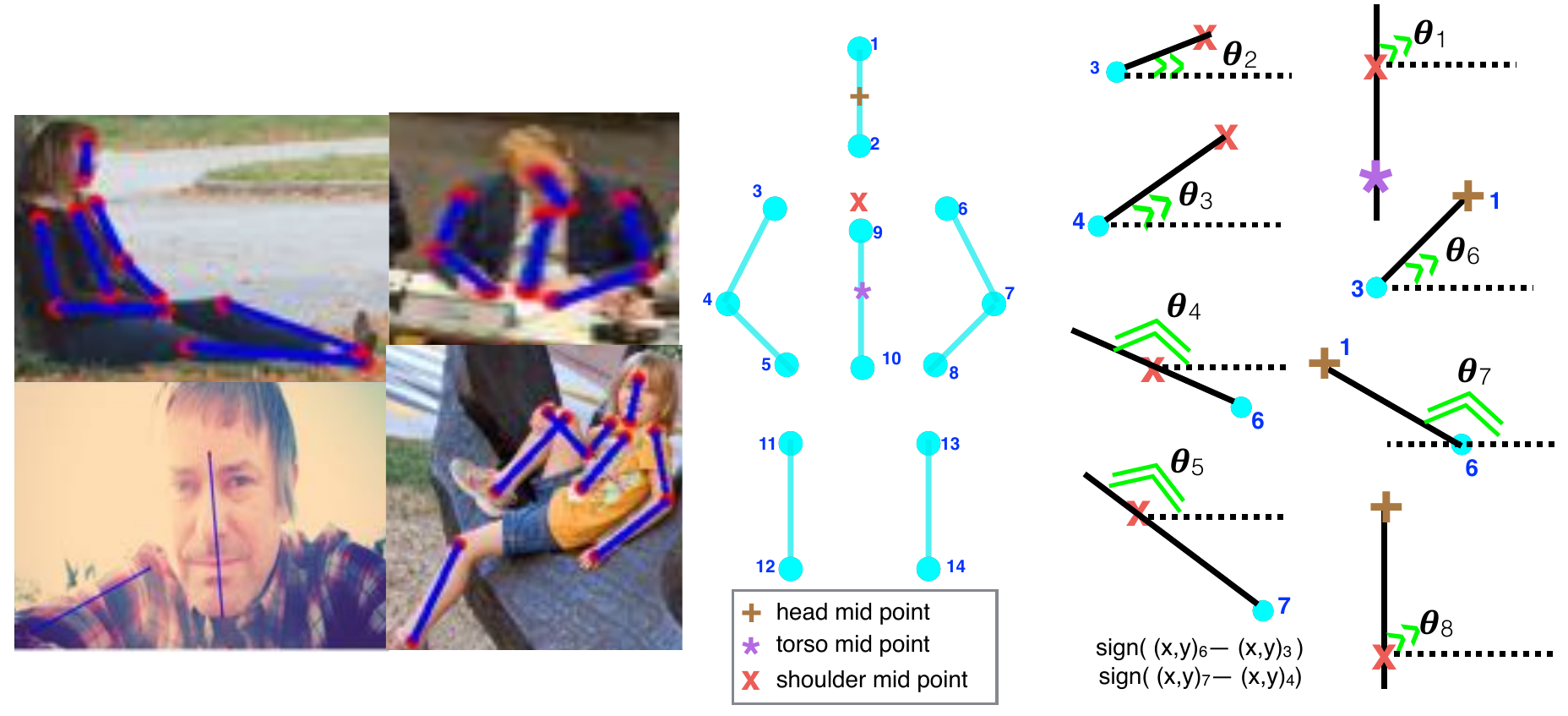}
\caption{We use body keypoints (left) to learn prominent views using a set of features (right) based on  orientation of informative body parts and keypoints. }
\label{fig:pose_annotation_samples}
\end{figure}

To facilitate pose-aware representations, we partition the training images into prominent views using body keypoints. Although people exhibit large variations in arm and leg positions, we consider only the informative regions such as head and torso. We construct a $24${-}$D$ feature for pose clustering using $14$ key points and visibility annotations as shown in Figure~\ref{fig:pose_annotation_samples}. It consists of -

\begin{enumerate}[nolistsep]
\item {\em $10${-}$D$ orientation feature} based on the relative location of different body parts computed as 
$ [cos(\theta_1),\dots,cos(\theta_8), sign(x_6-x_3), sign(x_7-x_4)]$, where $\theta_i, i=\{1,2,3,\dots,8\}$ denote the angle between the line joining two key points and the $x$-axis. For example, $\theta_6$ is the angle between head midpoint and right shoulder. The last two elements distinguish front and back views, and are based on the sign of $x$-coordinate differences of left and right points of shoulder and elbow, respectively.
\item {\em $14${-}$D$ visibility feature}, where each element is either $1$ or $0$ depending upon whether the corresponding keypoint is visible or not. This provides strong pose cues as certain body parts are not visible in particular views.
\end{enumerate}

To identify meaningful views, we apply k-means algorithm to cluster the images based on the above features. We first obtain a large number ($30$) of highly similar groups, which are then hierarchically merged to obtain seven prominent views. Figure~\ref{fig:pose_intro}(b) shows an example from each of these views for two different people. The views from left to right are: {\tt frontal}, {\tt semi-left}, {\tt left}, {\tt semi-right}, {\tt right}, {\tt back} and {\tt partial} views, respectively. The {\tt partial} cluster contains images where only the head and possibly part of shoulders are visible.

Once we obtain the prominent views or poses, we train an pose-estimator based on AlexNet that takes full body image as input and computes the likelihood of each pose. During testing, the pose likelihood estimated from the pose-estimator provide the pose-aware weights $w_i$, in Eqn.~\ref{eqn1}. We noticed from our experiments that, it is critical to $l_2$-normalize the weights to obtain the improved performance. 

\subsection{Learning a PSM}
\label{sec:psmlearn}
To train a pose-specific model ({\it PSM}), we select the training samples that belong to a specific pose cluster. We consider the head and upper body regions as these are the most informative cues for recognition \cite{person_recognition_bodyparts, person_recognition_context}. Given a head at location $(l_x, l_y)$ with dimensions $(\delta_x, \delta_y)$, we estimate the upper body to be a box at location $(l_x  - 0.5\alpha, l_y)$ of dimensions $(2\alpha, 4\alpha)$ where $\alpha = min(\delta_x, \delta_y)$.

Given different body parts, one possibility is to train independent convnets on each of these regions \cite{person_recognition_bodyparts,person_recognition_context,person_recognition_poselets}. However, discriminative body regions that help in recognition may vary across training instances. For example, Figure~\ref{fig:pose_intro}(a)-4 contains an occluded face region and is less informative. Similarly, upper body may be less informative in some other instances. If such noisy or less informative regions influence the optimization process, it may reduce the generalization ability of the networks.

\begin{figure}
\centering
\includegraphics[width = 0.95\linewidth]{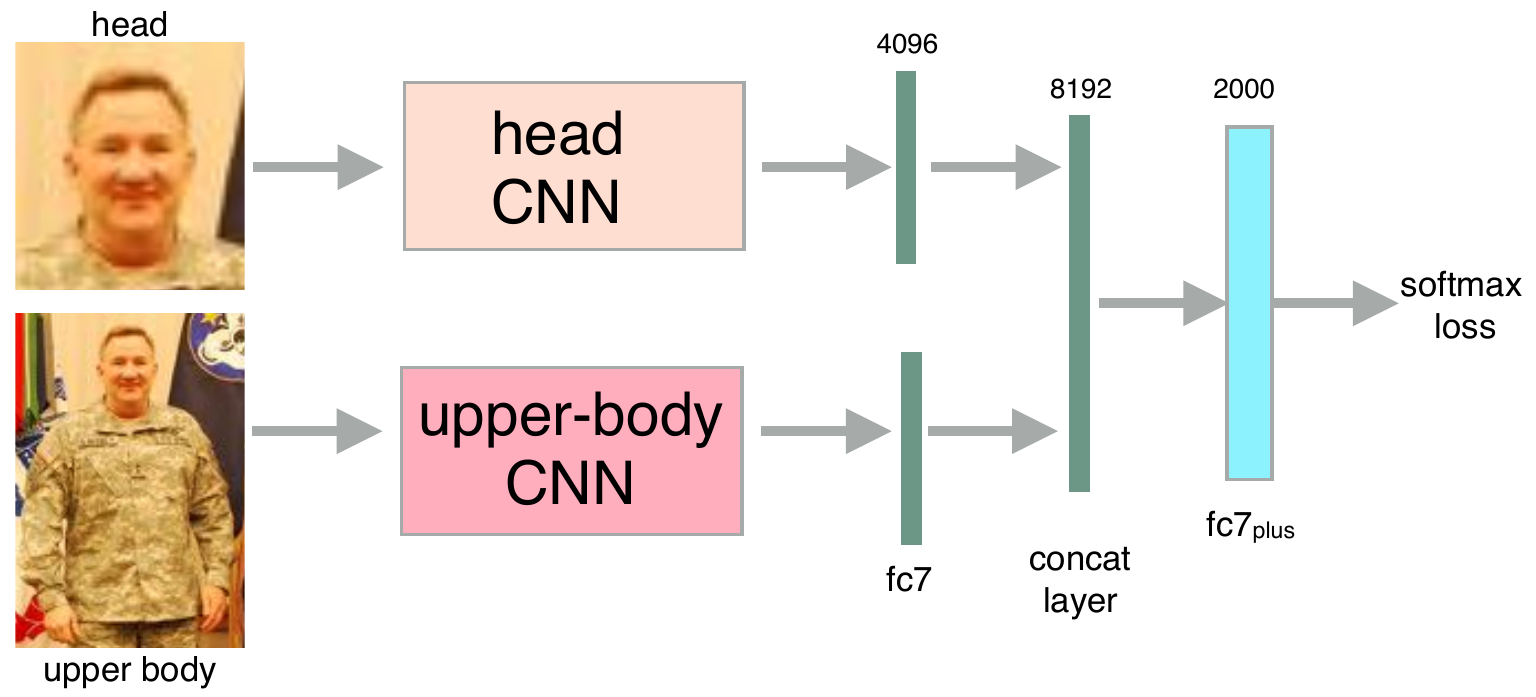}
\caption{{\bf PSM:} Our network consists of two AlexNets for head and upper body with a single output layer. The last fully connected layer of the two regions are concatenated and passed to an joint hidden layer with $2000$ nodes.}
\label{fig:joint_training}
\end{figure}

We propose an approach to improve the generalization ability by allowing the network to selectively focus on informative body regions during the training process. The idea is to optimize both the head and upper body networks jointly over a single loss function. Our {\it PSM} contains two AlexNets corresponding to the head and upper-body regions (see Figure~\ref{fig:joint_training}). The final {\tt fc7} layers of each region are concatenated and passed to a joint hidden layer ({\tt fc7}$_\text{plus}$) with $2000$ nodes before the classification layer. This provides more flexibility to the network to make the predictions based on one region even if the other region is noisy or less informative. As we show in our experiments (Table~\ref{table:joint_train_comparison}), the joint training approach performs better than separate training of regions. 

\subsection{Identity Prediction with PSMs}
\label{sec:predict}

We derive multiple features from each {\it PSM} and train classifiers on these feature vectors. The primary feature vector ($\mathcal{F}$) consists of
the sixth and seventh layers of the head ({\tt h}) and upper body ({\tt u}), and the joint fully connected layer ($\mathcal{F}{:}$ ${<}{\tt fc6_{\tt h}}, {\tt fc7_{\tt h}}, {\tt fc6_{\tt u}}, {\tt fc7_{\tt u}}, {\tt fc7}_{{\tt plus}}{>}$). In addition to $\mathcal{F}$, we define two additional feature vectors solely based on the head and upper body layers - \mbox{$\mathcal{F}_h{:}$ ${<}{\tt fc6_{\tt h}}, {\tt fc7_{\tt h}}{>}$} and $\mathcal{F}_u{:}$ ${<}{\tt fc6_{\tt u}}, {\tt fc7_{\tt u}}{>}$. We train linear SVM classifiers on each of the above feature vectors to obtain the identity predictions. The pose-specific identity score, $s_i(y,x)$, is simply the sum of the three SVM classifier outputs.
	\begin{equation} 
		s_i(y, x) = \sum_{f \in \{ \mathcal{F}, \mathcal{F}_h, \mathcal{F}_u\}} P_i(y|f; x),
	\label{eqn2}
	\end{equation}			
where $P_i(y|f;x)$ is the class $y$ score of the sample $x$ predicted by the classifier trained on the feature $f$ in $i$-th view.

\section{Experiments and Results}
\subsection{Datasets and Setup}
We select three datasets from the domain of photo-albums, movies and sport broadcast videos as shown in Figure~\ref{fig:datasets}. Each of these settings have their own set of advantages and challenges as summarized in Table~\ref{table:dataset_stats}. To the best of our knowledge, this is the first work that evaluates person recognition in such diverse scenarios. 

\subsubsection{Photo Album Dataset} \textsc{PIPA} \cite{person_recognition_poselets} consists of $37{,}107$ photos containing $63{,}188$ instances of $2{,}356$ identities collected from user-uploaded photos in Flickr. 
The dataset consists of four splits with an approximate ratio of $45$:$15$:$20$:$20$. The larger split is primarily used to train convnets, second split to optimize parameters during validation and the third split to evaluate recognition algorithms. 
The evaluation set is further divided into two equal subsets, each with $6{,}443$ instances belonging to $581$ subjects for training and testing the classifiers. We follow \textsc{PIPA} experimental protocol and train the classifiers on one fold and test on another fold, and vice-versa. 
We also conduct experiments on challenging splits introduced by \mbox{Oh {\em et al.}~\cite{person_recognition_bodyparts}} based on album, time and day information. 

\begin{figure}
\centering
\includegraphics[width = 0.98\linewidth]{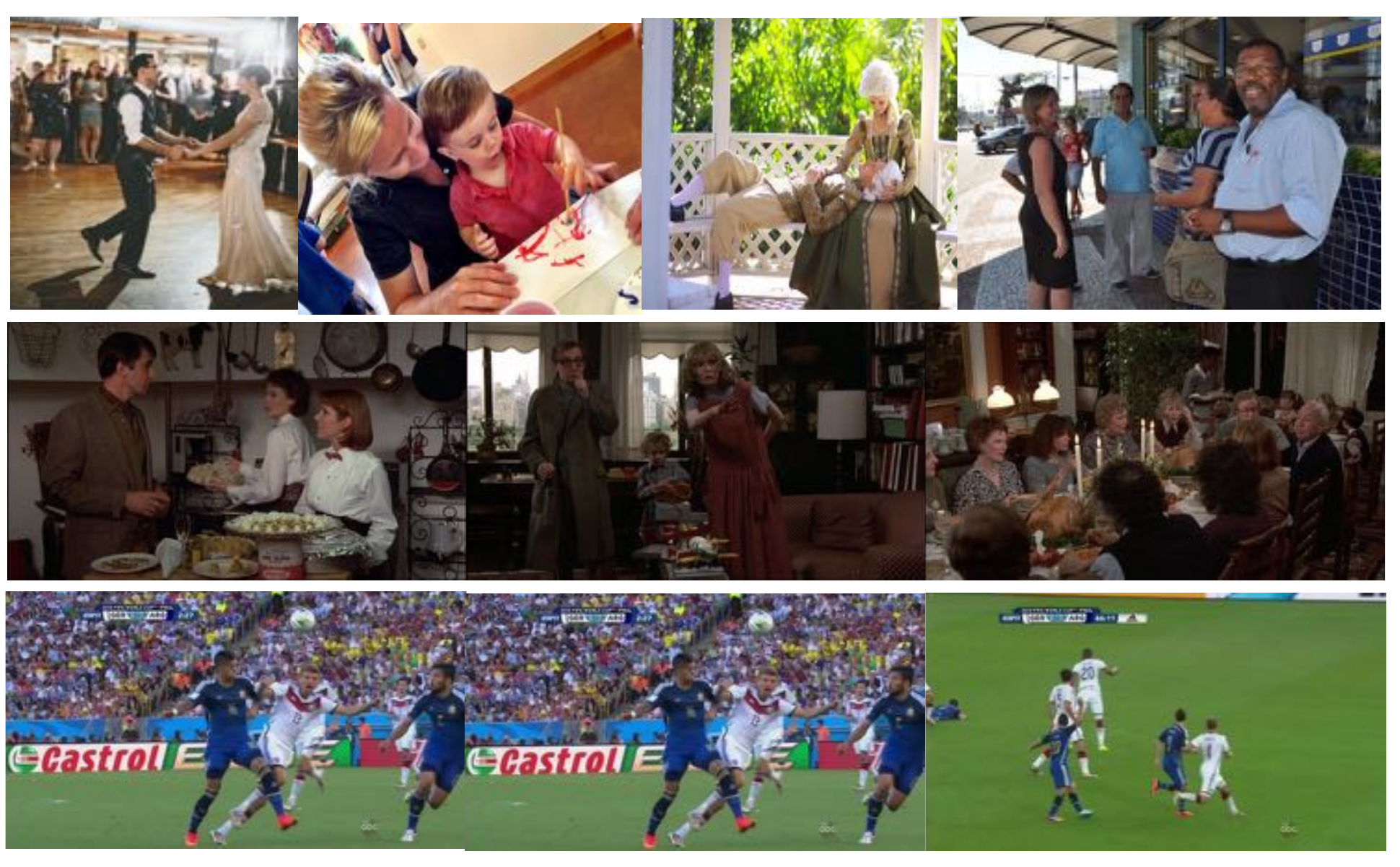}
\vspace*{-0.1in}
\caption{Few images from (top) PIPA, (middle) Hannah and (bottom) Soccer datasets.}
\label{fig:datasets}
\end{figure}

\subsubsection{Hannah Movie Dataset} We consider ``Hannah and Her Sisters'' dataset \cite{hannah} to recognize the actors appearing in the movie. 
The dataset consists of $153{,}833$ movie frames containing $245$ shots and $2{,}002$ tracks with $202{,}178$ face bounding boxes. We regress the face annotations to get the rough estimate of head. There are a total of $254$ labels of which $41$ are the named subjects. The remaining labels are the unnamed characters (boy1, girl1, \etc) and miscellaneous regions (crowd, \etc). 

To consider a more practical recognition setting, we create another dataset for classifier training using \textsc{IMDB} photos. For each named character, we collect photos from actor's profile in \textsc{IMDB}\cite{IMDB} and annotate the head bounding boxes. Of the $41$ named characters, only $26$ prominent actors had profiles in \textsc{IMDB}. The \textsc{IMDB} train set consists of $2{,}385$ images belonging to $26$ prominent actors appeared in the movie. 
There are a total of $159{,}458$ instances belonging to these $26$ actors in the test set. There is a significant age variation between train and test instances since the Hannah instances are created from a particular year ($1986$) while the \textsc{IMDB} photos are captured over a long period of time.  

\subsubsection{Soccer Dataset} We create soccer dataset from the broadcast video of World cup $2014$ final match played between Argentina and Germany. We considered only replay clips as these capture the important events of the match. We further filtered the replay clips to retain only those clips that are shot in close-up and medium views. 
We used \textsc{VATIC} toolbox \cite{vatic} to annotate the players in videos. Our soccer dataset consists of $37$ video clips with an average duration of $30$ secs. It consists of $28$ subjects with $13$ players from Germany team, $14$ players from Argentina team, and a referee.  

Unlike \textsc{PIPA}, we marked full-body bounding boxes for each player since head is not visible or out-of-view in many instances, and it also is difficult to estimate the bounding boxes of different body regions from head, due to large deformations. We followed \textsc{PIPA} annotation protocol and labeled the players regardless of their pose, resolution and visibility. We annotated the players to generate continuous tracks even in the presence of severe occlusion. Whenever it is difficult to recognize the players, we relied on additional clues such as hair, shoes, jersey number and accessories. However, we do not rely on any of these domain-specific cues in this work. For evaluation purposes, we randomly select $10$ clips into training and remaining $27$ clips into testing. This resulted in $19{,}813$ instances in training set and $51{,}051$ instances in testing set.

\begin{table}[!t]
  \centering
  \begin{tabular}{l|c|c|c}
         & PIPA \cite{person_recognition_poselets} & Hannah \cite{hannah} & Soccer\\ 
        \hline

        Train instances     & $6{,}443$ & $2{,}385$ & $19{,}813$\\
         \hline
	Train subjects   & $581$ &  $26$ & $28$\\
        \hline
        Test instances     & $6{,}443$ & $202{,}178$ & $51{,}051$\\        
	 \hline
	Test subjects   & $581$ &  $41$ & $28$\\
	\hline
	 Annotations   & Head &  Face & Body \\
	 \hline
	 Domain variation & No & Yes & No\\
	 \hline
	 Clothing &  Yes & No & No  \\
	 \hline
	 Age gap  & No & Yes & No\\
	\hline
	 Head resolution & High & Medium & Low\\	
	  \hline
	 Motion blur & No & Moderate & Severe\\	
	\hline
	 Deformation & Less & Moderate & Severe\\	
	\hline

\end{tabular}
\vspace*{-0.1in}
  \caption{Comparison of the datasets in terms of statistics, annotations, merits and challenges.}
  \label{table:dataset_stats}
\end{table}

\subsection{Results and Analysis}\label{sec:results_analysis}
\begin{figure}
\centering
\includegraphics[scale=0.5]{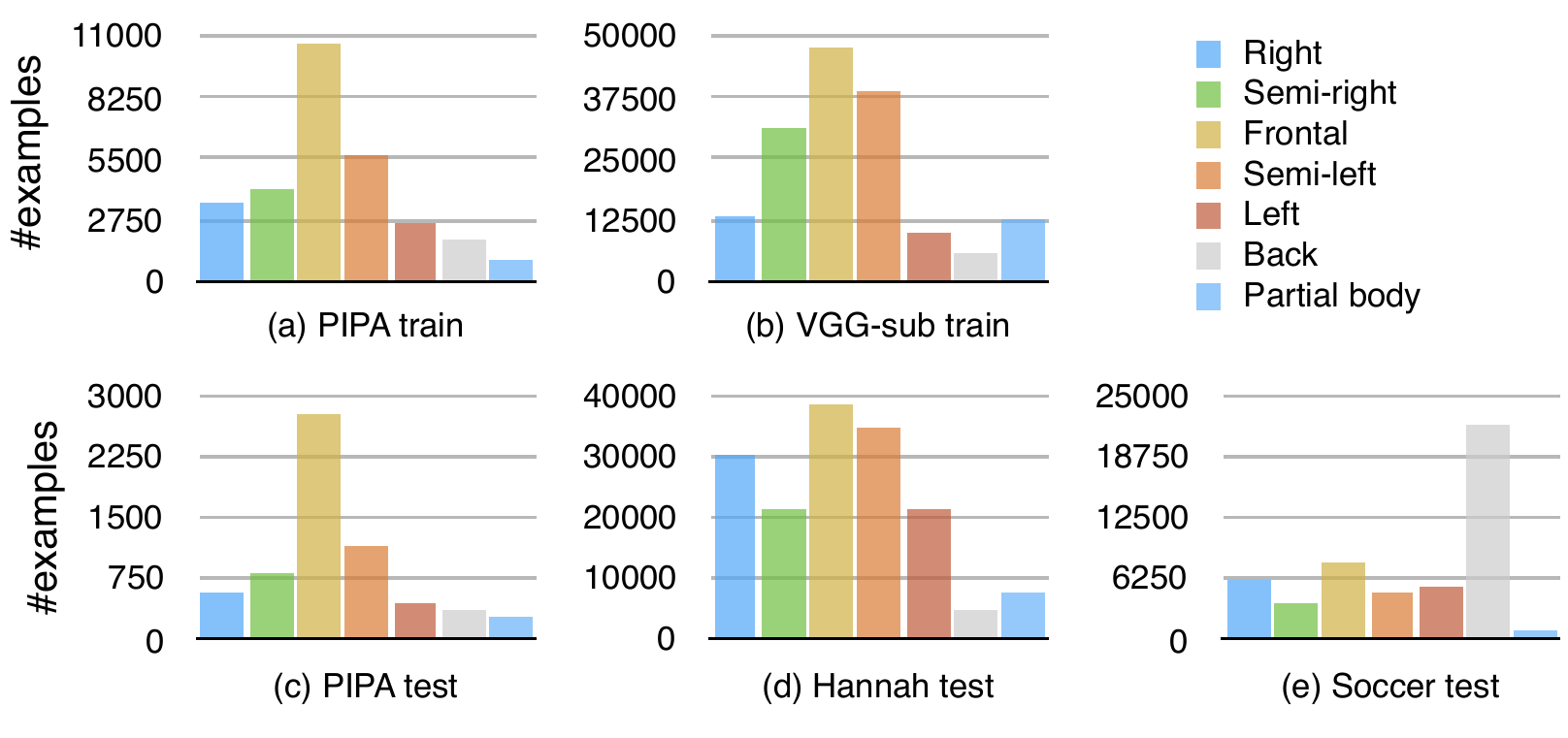}
\vspace*{-0.14in}
\caption{Pose statistics of different datasets.}
\label{fig:datasets_pose_stats}
\end{figure}

For all our experiments, we use the {\it PSM} models trained on larger set of \textsc{PIPA} consisting of $29{,}223$ instances. We annotate \textsc{PIPA} train instances with keypoint locations to learn prominent views as discussed in \autoref{sec:viewlearn}. The number of instances in each view after pose clustering is shown in Figure~\ref{fig:datasets_pose_stats}(a). We train a separate {\it PSM} on {\tt frontal}, {\tt semi-left}, {\tt left}, {\tt semi-right}, {\tt right}, {\tt back} and {\tt partial} views. The {\tt base} model is trained on the entire \textsc{PIPA} train set. We augment each view by horizontal flipping of the instances from its symmetrically opposite view. For instance, images in left view are flipped and augmented to right view. We use Caffe library \cite{caffe} for our implementation. For optimization, we use stochastic gradient descent with a batch size of $50$ and momentum coefficient of $0.9$. The learning rate is initially set to $0.001$, which is decreased by a factor of $10$ after every $50{,}000$ iterations. We train the networks for a total of $300{,}000$ iterations. The parameter $C$ is set to $1$ for training \textsc{SVM} and the base weight $w_0$ to $1$.

We noticed that the {\it PSM}s trained on view samples lead to over-fitting. To overcome this, we used a subset of VGGFace dataset~\cite{deepface_omkar} for initializing the networks. We extended the face annotations and selected only those instances that have full body in the \textsc{VGG} images. The number of examples used to initialize {\it PSM}s are shown in Figure~\ref{fig:datasets_pose_stats}(b). We make two important points regarding the additional data. First, the extra data ($\sim$$160$K) we considered is much smaller compared to {\tt PIPER} ($\sim$$4$M faces for training DeepFace~\cite{deepface}) and {\tt naeil} ($\sim$$500$K from four different datasets). Second, the proposed improvement is primarily due to the pose-aware combination strategy and not the ensemble of different view-specific models. This is discussed below in the Ablation study (III).

{\bf Overall performance:} 
We compare the performance of various approaches on \textsc{PIPA} test splits in Table~\ref{table:pipa_split_comaprison}, including both baseline and contextual results of Li \etal \cite{person_recognition_context} to provide a comprehensive review. When the contextual information is not considered, our approach outperforms all the previous approaches achieving an $89.05$\% on the original split. On the Hannah dataset, our approach outperforms {\tt naeil} by a large margin as shown in Table~\ref{table:hannah_performance}.  Finally, we show the results on the newly created player recognition in soccer in Table~\ref{table:soccer_performance}. 

We notice on all the datasets that use of multiple body regions helps in recognition strengthening the motivation behind {\tt PIPER} and {\tt naeil} algorithms. We also show the results by merging track labels on Hannah and Soccer datasets to understand their impact on recognition. We reassign the frame labels based on simple majority voting of all the frames in a track. The results suggest that track information if available should be used to improve the performance.

The accuracies on Hannah and Soccer datasets are much lower compared to \textsc{PIPA} owing to lower resolution, motion blur, heavy occlusions, and age variations. We also observe a little improvement over {\tt naeil} on soccer dataset due to unusual poses (kicking, falling, \etc.) of the subjects. A large majority of these images are predicted as ``back-view'' by the pose-estimator (see Figure~\ref{fig:datasets_pose_stats}(e)). Since the performance of our {\tt back} view model is poor due to limited training data, we noticed only minimal improvement. 

{\bf Ablation study ({\rm I}):} We analyze the effectiveness of different features and joint optimization strategy with {\tt base} model in Table~\ref{table:joint_train_comparison}. The use of both {\tt fc}$_6$ and {\tt fc}$_7$ features improve the performance for all the body regions. The head ($\mathcal{F}_h$) and upper body ($\mathcal{F}_u$) features obtained through joint training outperform the head ({\tt h}$_2$) and upper body ({\tt u}$_2$) features obtained through separate training by almost two percent points. Similarly, the concatenation of head and upper body features ($\mathcal{F}$) through joint training perform better than separate training ($\mathcal{S}_1$). Finally, the combination of three classifiers ($s_0(y,x)$) from head, upper-body and joint features further bring the performance improvement. We note that, our single {\tt base} model with joint-training strategy and combination of classifiers itself outperforms {\tt naeil}, which reports an accuracy of $86.78\%$ with $17$ models. 

\begin{figure*}[t]
\begin{center}
\subfigure{
 \includegraphics[width=0.35\linewidth]{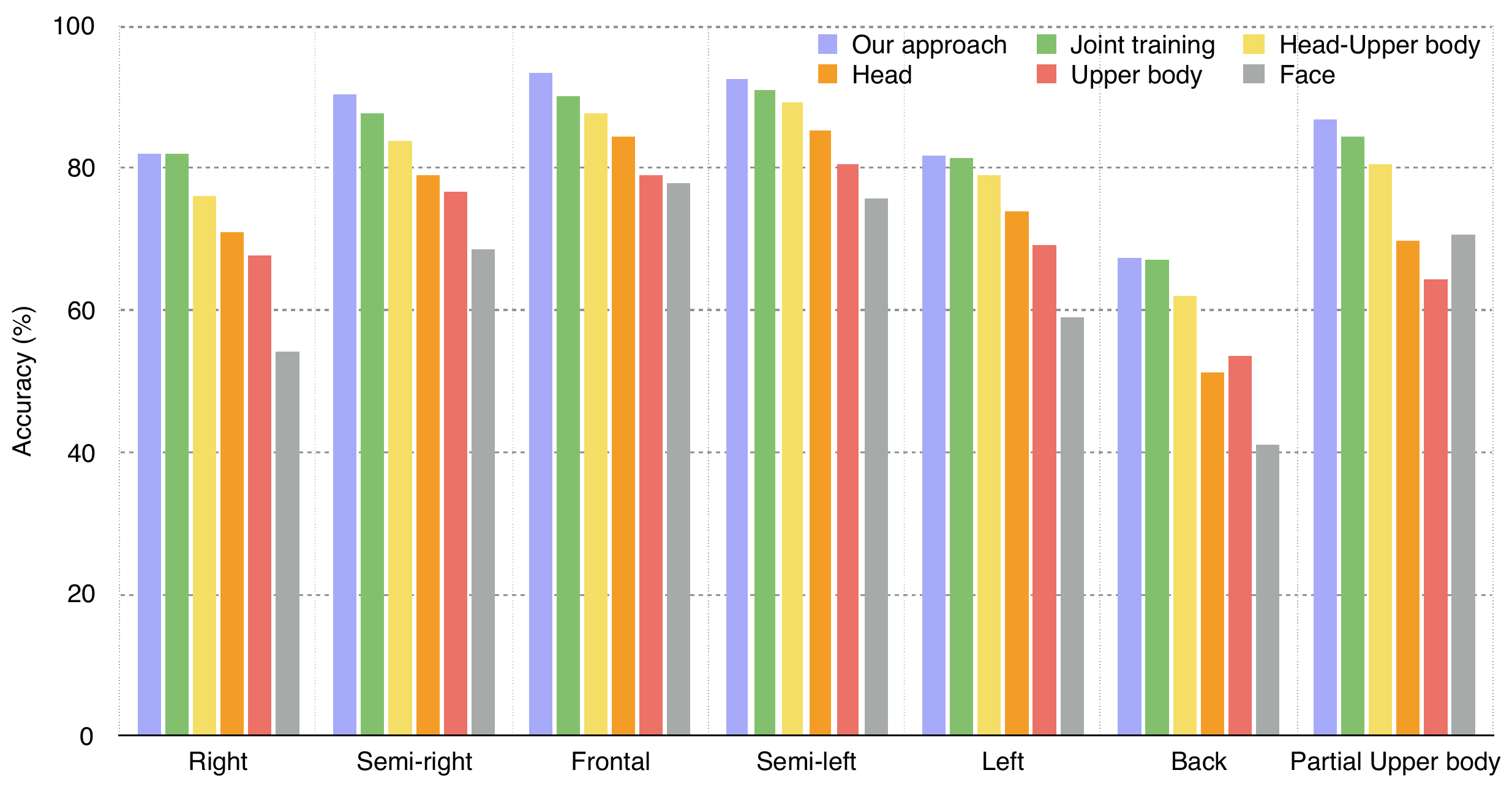}
}\subfigure{
 \includegraphics[width=0.315\linewidth]{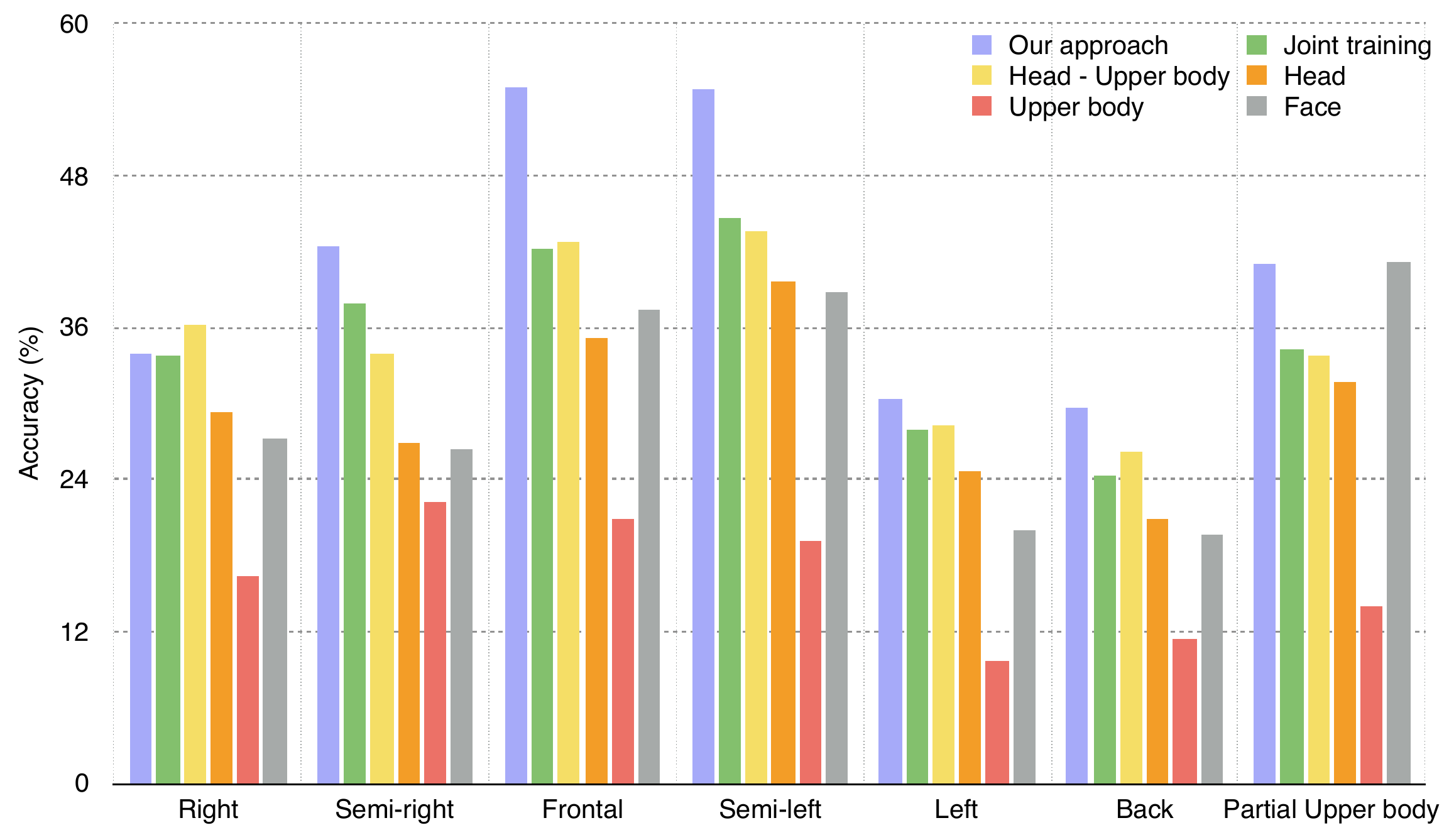}
}\subfigure{
 \includegraphics[width=0.293\linewidth]{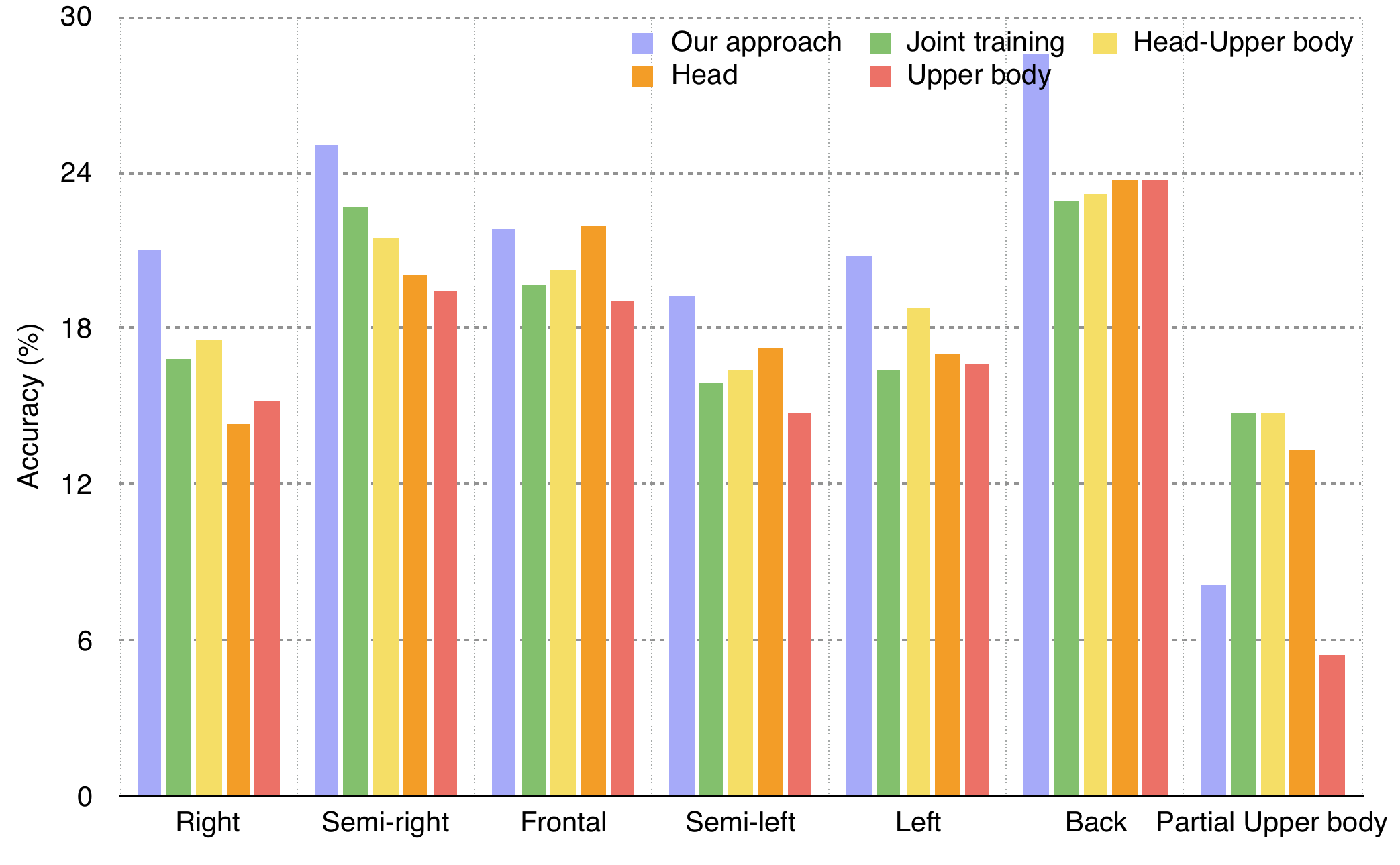}
}
\end{center}
\vspace*{-0.3in}
\caption{Pose-wise recognition performance on \textsc{PIPA} (left), Hannah (middle) and Soccer (right) datasets.}
\label{fig:per_pose_performance}
\end{figure*}

\begin{table}[t]
  \begin{center}
  \begin{tabular}{l|c|c|c|c}
         Method & Original & Album & Time & Day\\ 
        \hline
	{\tt PIPER}  \cite{person_recognition_poselets}        & 83.05& -&-&- \\		               	
	\hline	
	{\tt naeil} \cite{person_recognition_bodyparts}          & 86.78 & 78.72&69.29&46.61 \\
	\hline
	{\small Li \etal w/o context \cite{person_recognition_context}} & 83.86& 78.23&70.29&56.40\\
	\hline
	{\small Li \etal with context \cite{person_recognition_context}} & 88.75& {\bf 83.33}&{\bf77.00}&{\bf 59.35}\\
	\hline
	{\bf Our approach} & {\bf 89.05} & 82.37 & 74.84 & 56.73\\		                
	\hline
	\end{tabular}
      \end{center}
      \vspace*{-0.2in}
  \caption{Performance comparison (\%) of various approaches on different \textsc{PIPA} splits.}
  \label{table:pipa_split_comaprison}
\end{table}

\begin{table}[t]
  \begin{center}
  \begin{tabular}{l l l}
         \hline
         Method            & Accuracy & Accuracy\\
                                 & without tracks & with tracks \\ 
        \hline
	Head ({\tt H})           &	27.52 &31.91\\		               
	\hline
	Face ({\tt F})          &	26.53 &31.55\\	
	\hline
	Upper body ({\tt U}) &    16.49 &17.72\\		                
	\hline
	Separate training of {\tt H} and {\tt U}   &  31.86&36.10\\ 
	\hline
	Joint training of {\tt H} and {\tt U} &  32.92 & 37.74\\
	\hline
	{\tt naeil} \cite{person_recognition_bodyparts}&  31.41 & 37.57\\
	\hline
	{\bf Our approach}  & {\bf 40.95} & {\bf 44.46}\\ 
	\hline
\end{tabular}
\end{center}
\vspace*{-0.2in}
  \caption{Recognition performance (\%) of various approaches on Hannah movie dataset using \textsc{IMDB} dictionary.}
  \label{table:hannah_performance}
\end{table}

\begin{table}[t]
  \begin{center}
  \begin{tabular}{l l l}
         \hline
         Method            & Accuracy & Accuracy\\
                                 & without tracks & with tracks \\ 
        \hline
	Head ({\tt H})           &	17.68 &20.54\\		               
	\hline
	Upper body ({\tt U}) &   18.01 & 19.76\\		                
	\hline
	Separate training of {\tt H} and {\tt U}   &  17.62 & 20.68\\ 
	\hline
	Joint training of {\tt H} and {\tt U} &  18.35 & 20.18\\
	\hline
	{\tt naeil} \cite{person_recognition_bodyparts} &   19.45 &  23.77\\ 
	\hline
	{\bf Our approach}  &  {\bf 20.15} & {\bf 24.31}\\ 
	\hline
\end{tabular}
\end{center}
\vspace*{-0.2in}
  \caption{Performance comparison (\%) on Soccer dataset.}
  \label{table:soccer_performance}
\end{table}

\begin{table}[!t]
  \begin{center}
    \begin{tabular}{>{\arraybackslash}m{0.95in} | >{\arraybackslash}m{1.5in} | >{\arraybackslash}m{0.47in} }
                   & Feature  & Accuracy\\ 
        \hline
        Face ({\tt F})          &      {\tt fc7}$_\text{f}$  &	66.83 \\
		               &      [{\tt fc6}$_\text{f}$  {\tt fc7}$_\text{f}$]  &    70.40 \\
         \hline
	Head ({\tt H})          &      {\tt fc7}$_\text{h}$~~~~~~~~~~~~~~~~~~~~~~~~\dots({\tt h}$_\text{1}$) &	76.81 \\
		                &      [{\tt fc6}$_\text{h}$  {\tt fc7}$_\text{h}$]~~~~~~~~~~~~\dots({\tt h}$_\text{2}$)  &    79.54 \\
	\hline
	Upper body ({\tt U}) &      {\tt fc7}$_\text{u}$ ~~~~~~~~~~~~~~~~~~~~~~~\dots({\tt u}$_\text{1}$)   &	72.26 \\
		                 &      [{\tt fc6}$_\text{u}$  {\tt fc7}$_\text{u}$]~~~~~~~~~~~~\dots({\tt u}$_\text{2}$)  & 	 75.19 \\
	\hline
	Separate training  &  [{\tt h}$_\text{1}$   {\tt u}$_\text{1}$]  &	82.90 \\
	of {\tt H} and {\tt U}                &  [{\tt h}$_\text{2}$  {\tt u}$_\text{2}$] ~~~~~~~~~~~~~~~~~~~~\dots($\mathcal{S}_1$)   & 84.01\\                  
	\hline
	  			&  {\tt fc7}$_\text{plus}$  &	85.98 \\
				&   [{\tt fc6}$_\text{h}$  {\tt fc7}$_\text{h}$] ~~~~~~~~~~~\dots($\mathcal{F}_h$)& 82.22\\
	Joint training	&   [{\tt fc6}$_\text{u}$  {\tt fc7}$_\text{u}$] ~~~~~~~~~~~\dots($\mathcal{F}_u$)& 77.62\\
	of {\tt H} and {\tt U}      &  [{\tt fc7}$_\text{h}$  {\tt fc7}$_\text{u}$]   & 85.05 \\
	        			&   [{\tt j}$_\text{1}$  {\tt fc7}$_\text{h}$  {\tt fc7}$_\text{u}$]  &85.22\\
			       & [{\tt fc7}$_\text{plus}$  {\tt fc6}$_\text{h}$  {\tt fc6}$_\text{u}$] &86.10\\
			       & [{\tt fc7}$_\text{plus}$ $\mathcal{F}_h$ $\mathcal{F}_u$]~~~~~~~\dots($\mathcal{F}$) & 86.27\\       
       			        &$s_0(y,x)$ & 86.96\\ 	
	\hline
\end{tabular}
  \end{center}
        \vspace*{-0.2in}
  \caption{Performance (\%) of different features obtained from separate and joint training of regions on \textsc{PIPA} test set.}
  \label{table:joint_train_comparison}
\end{table}

\begin{figure}
\centering
\includegraphics[width = 0.7\linewidth]{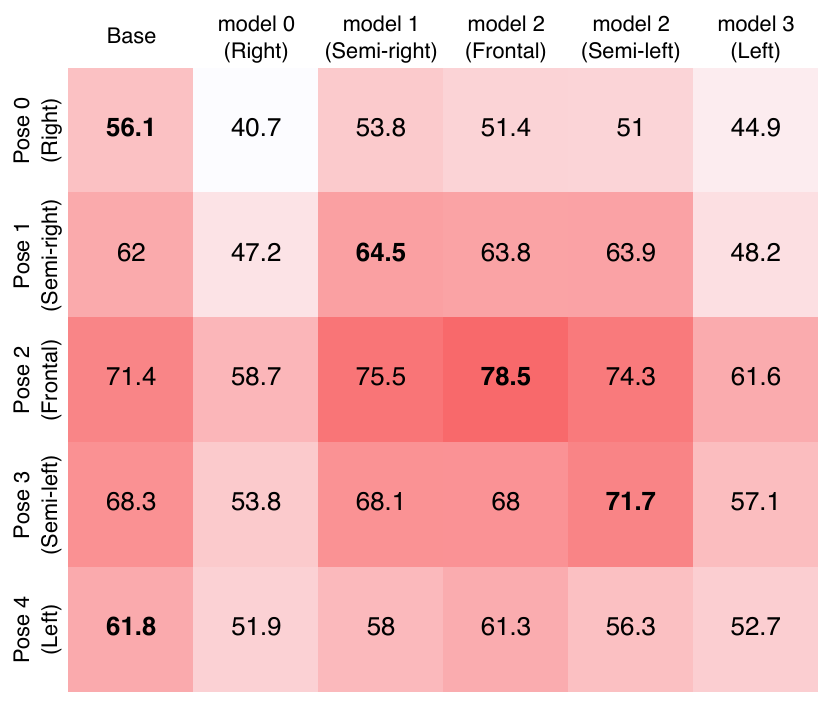}
\vspace*{-0.15in}
\caption{Effectiveness of {\it PSM}s: Each row shows the performance of test examples in a particular pose represented using the different {\it PSM}s.} 
\label{fig:pipa_performance_matrix}
\end{figure}

\begin{table}[!t]
  \begin{center}
  \begin{tabular}{>{\arraybackslash}m{1.1in} | >{\centering\arraybackslash}m{0.7in} ||| >{\centering\arraybackslash}m{0.7in} }
         Fusion type   & (I) & (II) \\
        \hline
	Average pooling    &83.57\%   &87.78\% \\		               
	\hline
	Max pooling &   80.54\% & 85.51\% \\	
	\hline
	Elementwise multiplication &   81.71\% & 86.32\% \\		                
	\hline
	Concatenation   &  {\bf 84.44\%} & 87.62\% \\ 
	\hline
	Pose-aware weights &  -- & {\bf 89.05\%} \\	
	\hline
\end{tabular}
\end{center}
\vspace*{-0.2in}
  \caption{Comparison of different fusion schemes for combining (i) features during joint training (using {\tt frontal} {\it PSM}) and (ii) pose-aware classifier scores during testing.}
  \label{table:pooling}
\end{table}


\begin{figure*}
\centering
\includegraphics[width = 0.95\linewidth]{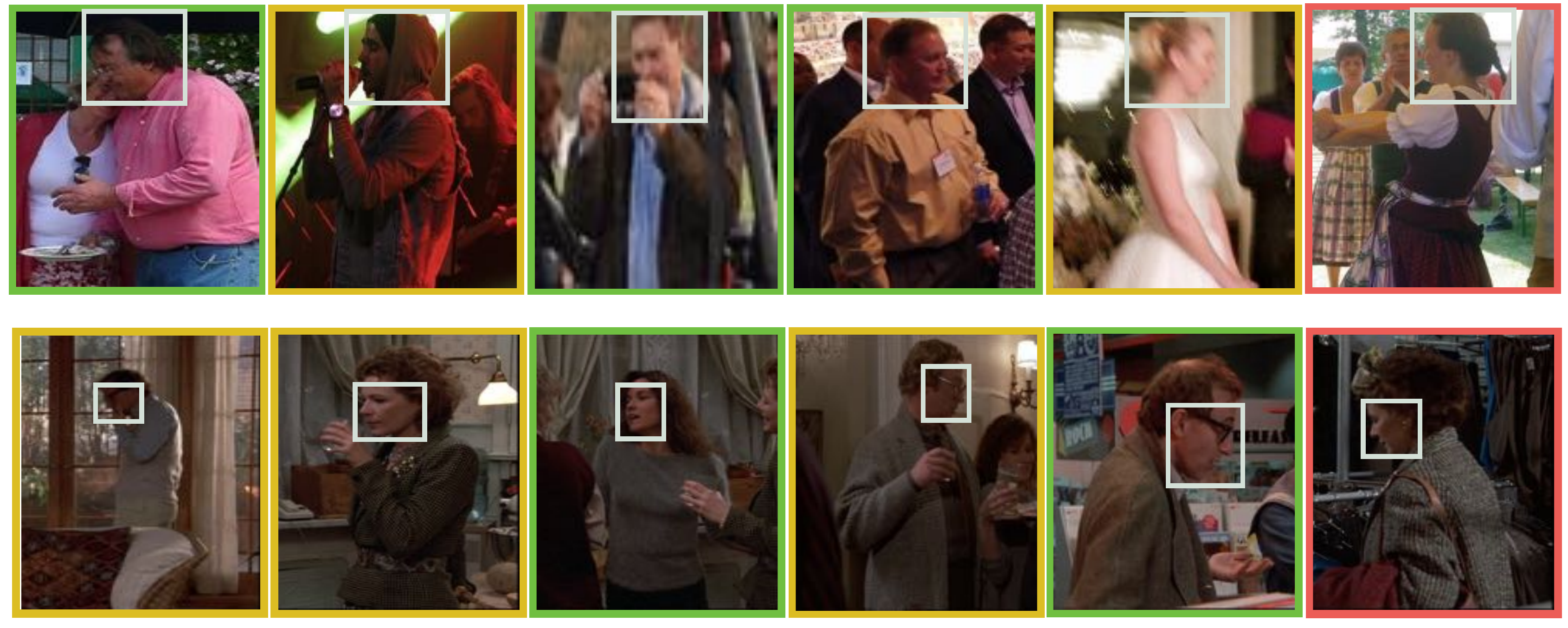}
\vspace*{-0.1in}
\caption{Success and failure cases on (top) \textsc{PIPA} and (bottom) Hannah. First five columns show the success cases of our approach where the improvement is primarily due to the specific-pose model. Green and yellow boxes indicate the success and failure result of {\tt naeil} respectively. Last column in red shows the failure cases for both our approach and {\tt naeil}.}
\label{fig:datasets}
\end{figure*}

\begin{figure}
\includegraphics[scale=0.32]{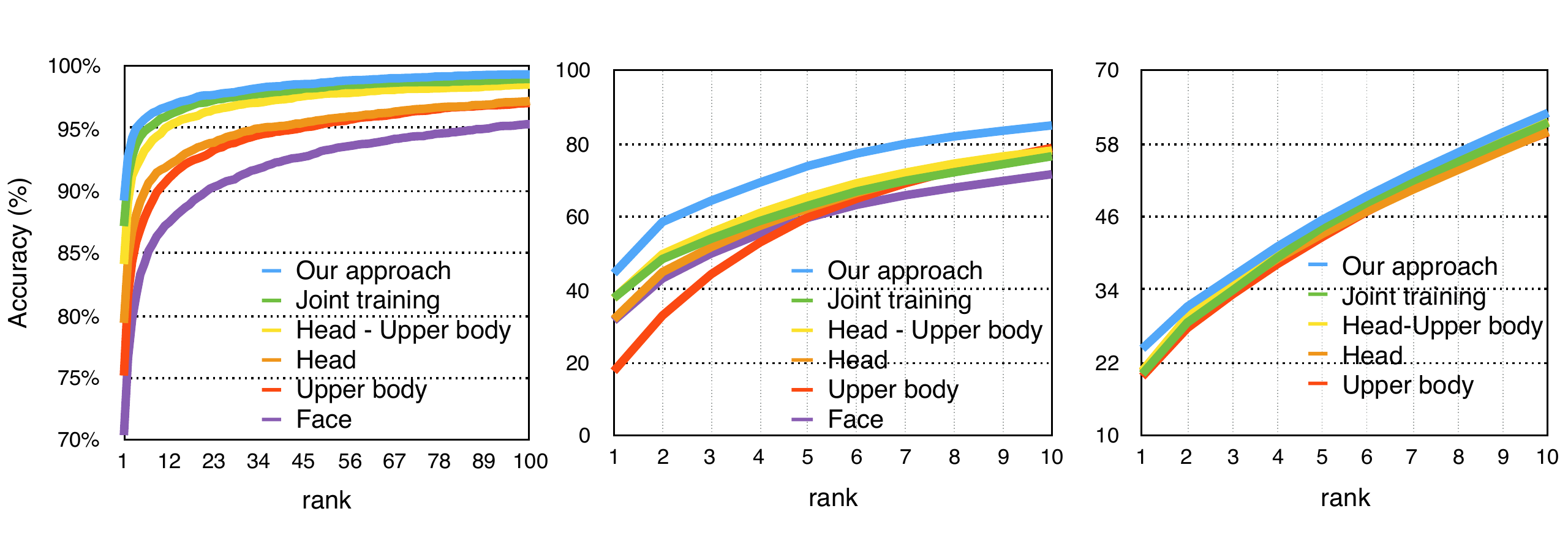}
\vspace*{-0.3in}
\caption{\textsc{CMC} curves of various approaches on \textsc{PIPA} (left), Hannah (middle) and Soccer (right) datasets.}
\label{fig:rankn}
\end{figure}

\begin{SCfigure}
\includegraphics[scale=0.31]{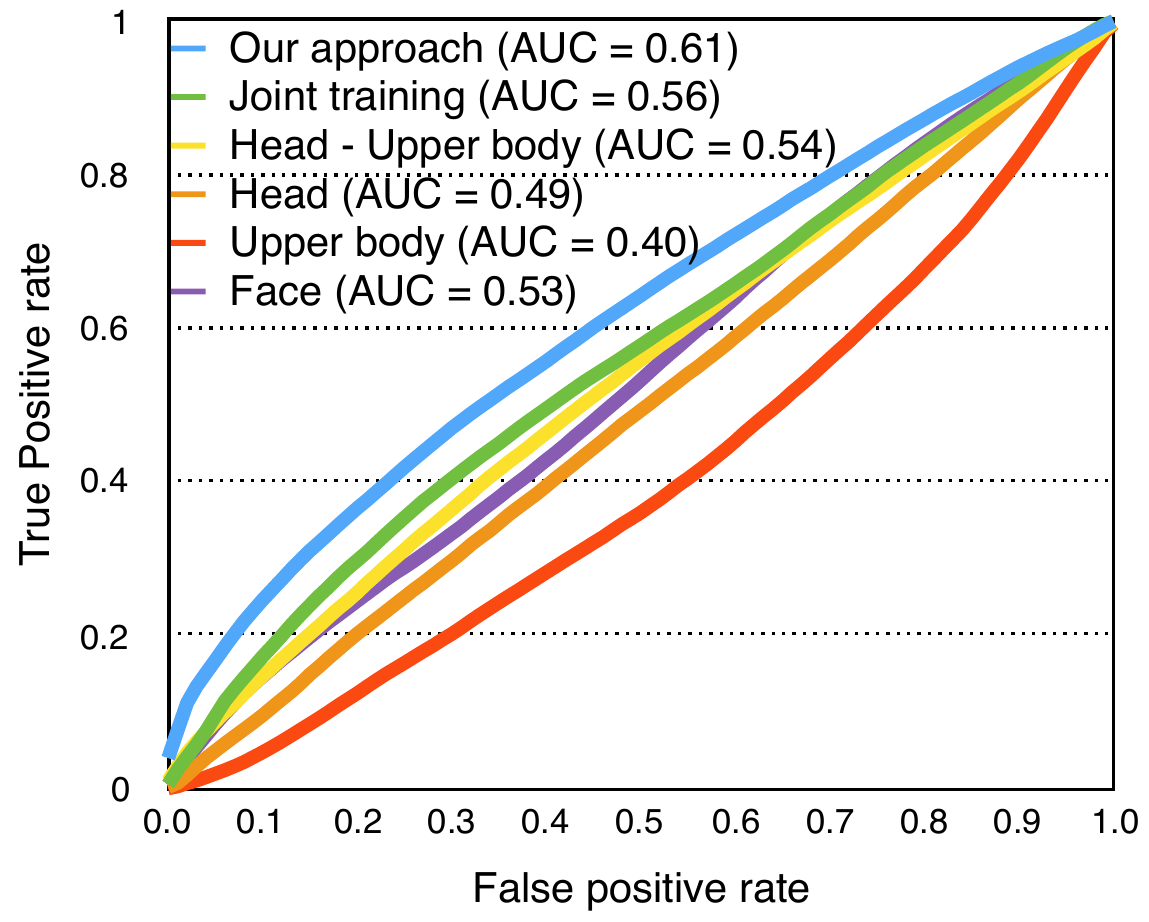}
\vspace*{-0.08in}
\caption{\textsc{ROC} curves of various approaches in rejecting unknown Hannah instances based on normalized prediction scores.}
\label{fig:hannah_roc}
\end{SCfigure}

{\bf Ablation study ({\rm II}):} We conduct experiments to measure the performance of pose-specific {\it PSM} models using \textsc{PIPA} test set. In each experiment, we consider only those examples that are in $i$-th pose and select half of them randomly for classifier training and remaining half into testing. We extract $\mathcal{F}_u$ feature for these examples using {\it PSM} and {\tt base} models. Figure~\ref{fig:pipa_performance_matrix} shows the performance of each model in recognizing examples from different views. It shows that for {\tt frontal}, {\tt semi-left} and {\tt semi-right} examples, corresponding {\it PSM} models outperform other models including the base model. This show that the person representations obtained from pose-specifc models are more robust than the pose-agnostic representations. However, for extreme profile views, we noticed that base model performed better than the corresponding profile models. We attribute this to the non-availability of enough profile images while training the {\it PSM}. For this reason we include the base model along with {\it PSM}s to bring more robustness when handling rear and non-prominent view images.

{\bf Ablation study ({\rm III}):} Our approach uses two kinds of information pooling, one during {\it PSM} training and another for combining classifiers. For joint-training, we show the effect of different head ({\tt fc7}$_\text{h}$) and upper body ({\tt fc7}$_\text{u}$) combination strategies in Table~\ref{table:pooling} (I). The simple concatenation of {\tt fc7}$_\text{h}$ and {\tt fc7}$_\text{u}$ worked better, and hence considered. Similarly, we tried multiple strategies to combine the classifiers during testing. As can be seen in  Table~\ref{table:pooling} (II), pose-aware weighting outperform other strategies including average pooling of the ensemble of classifiers.

{\bf Pose-wise recognition performance:} The statistics of different poses is given in Figure~\ref{fig:datasets_pose_stats} (bottom) for different datasets. Frontal images dominate \textsc{PIPA} due to which algorithms already achieve high performance ($>$$80$\%). Hannah consists of different poses in similar proportion while the soccer dataset contains majority of back view images. Consequently, we observe a low performance on these datasets. In Figure~\ref{fig:per_pose_performance}, we show pose-wise recognition performance. \textsc{PIPA} and Hannah have a similar trend in which frontal and semi-profile images are recognized with greater accuracies, while profile and back-views with less accuracy. The upper body seems to be less informative in case of Hannah as the clothing is completely different between Hannah and \textsc{IMDB}. The proportion of back views that are correctly recognized is slightly better in soccer setting due to large number of back view images in the classifier train set. 

{\bf Rank-n identification rates:} The Cumulative matching Characteristic (\textsc{CMC}) curves are shown in Fig~\ref{fig:rankn}. Our approach achieves rank-$10$ accuracies of $96.56$\%, $84.83$\% and $63.2$\% on \textsc{PIPA}, Hannah and Soccer datasets respectively. On Hannah, we noticed a big difference of $12$\% between rank-$1$ and rank-$2$ performance. The performance gap between different approaches tend to reduce with higher rank. 

{\bf Handling unknown instances:} In the movie scenario, the test set has $41$ ground truth labels while there are only $26$ subjects in the trainset. Therefore, recognition algorithms should have the ability to reject such unknown instances. To achieve this, we $l_2$-normalized the predicted class scores and considered the maximum score as a confidence measure. The confidence score obtained on pose-aware representations are more robust in rejecting unknown, the performance being measured using \textsc{ROC} curve in Figure~\ref{fig:hannah_roc}. 

{\bf Computational complexity:} The number of features extracted from each {\it PSM} is $18{,}384$ ($4096\times4 + 2000$). With $7$ pose-aware models and a base model, our total feature dimension is ($18,384 \times 8$) which is $\sim$$3$ times smaller than {\tt PIPER} ($4096 \times 109$) and $\sim$$2$ times larger than {\tt naeil} ($4096\times17$). For memory critical applications, {\tt fc7}$_\text{plus}$ alone can be used as feature. We achieve an accuracy of $87.01$\% on \textsc{PIPA} with {\tt fc7}$_\text{plus}$ still outperforming {\tt naeil} with a feature dimension of just $16{,}000$ ($2000\times 8$).
 
 \section{Conclusion}
 We show that learning a pose-specific person representation helps to better capture the discriminative features in different poses. A pose-aware fusion strategy is proposed to combine the classifiers using weights obtained from a pose estimator. The person representations obtained using a joint optimization strategy is shown to be more powerful compared to separate training of body regions. We achieve state-of-the-art results on three different datasets from photo-albums, movies and sport domains.

{\small
\bibliographystyle{ieee}
\bibliography{egbib}

\begin{thebibliography}{10}\itemsep=-1pt

\bibitem{IMDB}
Hannah and her sisters (1986), full cast and crew.
\newblock \url{http://www.imdb.com/title/tt0091167/fullcredits}.

\bibitem{pose_aware_face02}
W.~AbdAlmageed, Y.~Wu, S.~Rawls, S.~Harel, T.~Hassner, I.~Masi, J.~Choi,
  J.~Lekust, J.~Kim, P.~Natarajan, et~al.
\newblock Face recognition using deep multi-pose representations.
\newblock In {\em WACV}, 2016.

\bibitem{person_reid11}
E.~Ahmed, M.~Jones, and T.~K. Marks.
\newblock An improved deep learning architecture for person re-identification.
\newblock In {\em CVPR}, 2015.

\bibitem{lbp_face}
T.~Ahonen, A.~Hadid, and M.~Pietikainen.
\newblock Face description with local binary patterns: Application to face
  recognition.
\newblock {\em PAMI}, 2006.

\bibitem{face_recognition_context01}
D.~Anguelov, K.-c. Lee, S.~B. Gokturk, and B.~Sumengen.
\newblock Contextual identity recognition in personal photo albums.
\newblock In {\em CVPR}, 2007.

\bibitem{face_alignment01}
T.~Berg and P.~Belhumeur.
\newblock Poof: Part-based one-vs.-one features for fine-grained
  categorization, face verification, and attribute estimation.
\newblock In {\em CVPR}, 2013.

\bibitem{soccer01}
M.~Bertini, A.~Del~Bimbo, and W.~Nunziati.
\newblock Player identification in soccer videos.
\newblock In {\em SIGMM workshop on Multimedia information retrieval}, 2005.

\bibitem{poselets}
L.~Bourdev and J.~Malik.
\newblock Poselets: Body part detectors trained using 3d human pose
  annotations.
\newblock In {\em ICCV}, 2009.

\bibitem{CACD}
B.-C. Chen, C.-S. Chen, and W.~H. Hsu.
\newblock Cross-age reference coding for age-invariant face recognition and
  retrieval.
\newblock In {\em European Conference on Computer Vision}, 2014.

\bibitem{face_alignment05}
G.~Ch{\'e}ron, I.~Laptev, and C.~Schmid.
\newblock {P-CNN: Pose-based CNN Features for Action Recognition}.
\newblock In {\em ICCV}, 2015.

\bibitem{PETA}
Y.~Deng, P.~Luo, C.~C. Loy, and X.~Tang.
\newblock Pedestrian attribute recognition at far distance.
\newblock In {\em MM}, 2014.

\bibitem{recognition_video02}
M.~Everingham, J.~Sivic, and A.~Zisserman.
\newblock Taking the bite out of automated naming of characters in tv video.
\newblock {\em Image and Vision Computing}, 2009.

\bibitem{person_reid05}
M.~Farenzena, L.~Bazzani, A.~Perina, V.~Murino, and M.~Cristani.
\newblock Person re-identification by symmetry-driven accumulation of local
  features.
\newblock In {\em CVPR}, 2010.

\bibitem{recognition_video01}
V.~Gandhi and R.~Ronfard.
\newblock Detecting and naming actors in movies using generative appearance
  models.
\newblock In {\em CVPR}, 2013.

\bibitem{face_recognition_context04}
R.~Garg, S.~M. Seitz, D.~Ramanan, and N.~Snavely.
\newblock Where's waldo: Matching people in images of crowds.
\newblock In {\em CVPR}, 2011.

\bibitem{person_reid00}
S.~Gong, M.~Cristani, S.~Yan, and C.~C. Loy.
\newblock {\em Person re-identification}.
\newblock Springer, 2014.

\bibitem{sift_face2}
M.~Guillaumin, J.~Verbeek, and C.~Schmid.
\newblock Is that you? metric learning approaches for face identification.
\newblock In {\em ICCV}, 2009.

\bibitem{face_alignment02}
T.~Hassner, S.~Harel, E.~Paz, and R.~Enbar.
\newblock Effective face frontalization in unconstrained images.
\newblock In {\em CVPR}, 2015.

\bibitem{person_reid03}
M.~Hirzer, P.~M. Roth, M.~K{\"o}stinger, and H.~Bischof.
\newblock Relaxed pairwise learned metric for person re-identification.
\newblock In {\em ECCV}, 2012.

\bibitem{lfw}
G.~B. Huang, M.~Ramesh, T.~Berg, and E.~Learned-Miller.
\newblock Labeled faces in the wild: A database for studying face recognition
  in unconstrained environments.
\newblock Technical report, University of Massachusetts, Amherst, 2007.

\bibitem{caffe}
Y.~Jia, E.~Shelhamer, J.~Donahue, S.~Karayev, J.~Long, R.~Girshick,
  S.~Guadarrama, and T.~Darrell.
\newblock Caffe: Convolutional architecture for fast feature embedding.
\newblock {\em arXiv preprint arXiv:1408.5093}, 2014.

\bibitem{pose_aware_face03}
S.~Johnson and M.~Everingham.
\newblock Clustered pose and nonlinear appearance models for human pose
  estimation.
\newblock In {\em BMVC}, 2010.

\bibitem{person_recognition_bodyparts}
S.~Joon~Oh, R.~Benenson, M.~Fritz, and B.~Schiele.
\newblock Person recognition in personal photo collections.
\newblock In {\em {ICCV}}, 2015.

\bibitem{alexnet}
A.~Krizhevsky, I.~Sutskever, and G.~E. Hinton.
\newblock Imagenet classification with deep convolutional neural networks.
\newblock In {\em NIPS}, 2012.

\bibitem{lfw_survey}
E.~Learned-Miller, G.~B. Huang, A.~RoyChowdhury, H.~Li, and G.~Hua.
\newblock Labeled faces in the wild: A survey.
\newblock In {\em Advances in Face Detection and Facial Image Analysis}, 2016.

\bibitem{person_recognition_context}
H.~Li, J.~Brandt, Z.~Lin, X.~Shen, and G.~Hua.
\newblock A multi-level contextual model for person recognition in photo
  albums.
\newblock In {\em {CVPR}}, 2016.

\bibitem{person_reid09}
W.~Li, R.~Zhao, T.~Xiao, and X.~Wang.
\newblock Deepreid: Deep filter pairing neural network for person
  re-identification.
\newblock In {\em CVPR}, 2014.

\bibitem{person_reid06}
C.~Liu, S.~Gong, C.~C. Loy, and X.~Lin.
\newblock Person re-identification: What features are important?
\newblock In {\em ECCV}, 2012.

\bibitem{soccer02}
W.-L. Lu, J.-A. Ting, J.~J. Little, and K.~P. Murphy.
\newblock Learning to track and identify players from broadcast sports videos.
\newblock {\em PAMI}, 2013.

\bibitem{person_reid08}
B.~Ma, Y.~Su, and F.~Jurie.
\newblock Local descriptors encoded by fisher vectors for person
  re-identification.
\newblock In {\em ECCV}, 2012.

\bibitem{pose_aware_face01}
I.~Masi, S.~Rawls, G.~Medioni, and P.~Natarajan.
\newblock Pose-aware face recognition in the wild.
\newblock In {\em CVPR}, 2016.

\bibitem{hannah}
A.~Ozerov, J.-R. Vigouroux, L.~Chevallier, and P.~P{\'e}rez.
\newblock On evaluating face tracks in movies.
\newblock In {\em ICIP}, 2013.

\bibitem{deepface_omkar}
O.~M. Parkhi, A.~Vedaldi, and A.~Zisserman.
\newblock Deep face recognition.
\newblock {\em BMVC}, 2015.

\bibitem{gabor_face}
N.~Pinto, J.~J. DiCarlo, and D.~D. Cox.
\newblock How far can you get with a modern face recognition test set using
  only simple features?
\newblock In {\em CVPR}, 2009.

\bibitem{facenet}
F.~Schroff, D.~Kalenichenko, and J.~Philbin.
\newblock Facenet: A unified embedding for face recognition and clustering.
\newblock In {\em {CVPR}}, 2015.

\bibitem{fisher_face}
K.~Simonyan, O.~M. Parkhi, A.~Vedaldi, and A.~Zisserman.
\newblock Fisher vector faces in the wild.
\newblock In {\em BMVC}, 2013.

\bibitem{recognition_video03}
J.~Sivic, M.~Everingham, and A.~Zisserman.
\newblock ``{W}ho are you?'' {L}earning person specific classifiers from video.
\newblock In {\em CVPR}. IEEE, 2009.

\bibitem{face_recognition_context03}
J.~Sivic, C.~L. Zitnick, and R.~Szeliski.
\newblock Finding people in repeated shots of the same scene.
\newblock In {\em BMVC}, volume~2, page~3, 2006.

\bibitem{deepID03}
Y.~Sun, X.~Wang, and X.~Tang.
\newblock Deep learning face representation from predicting 10,000 classes.
\newblock In {\em {CVPR}}, pages 1891--1898, 2014.

\bibitem{deepface}
Y.~Taigman, M.~Yang, M.~Ranzato, and L.~Wolf.
\newblock Deepface: Closing the gap to human-level performance in face
  verification.
\newblock In {\em {CVPR}}, 2014.

\bibitem{recognition_video04}
M.~Tapaswi, M.~B{\"a}uml, and R.~Stiefelhagen.
\newblock âknock! knock! who is it?" probabilistic person identification in
  tv-series.
\newblock In {\em CVPR}, 2012.

\bibitem{person_reid12}
E.~Ustinova, Y.~Ganin, and V.~Lempitsky.
\newblock Multiregion bilinear convolutional neural networks for person
  re-identification.
\newblock {\em arXiv preprint arXiv:1512.05300}, 2015.

\bibitem{vatic}
C.~Vondrick, D.~Patterson, and D.~Ramanan.
\newblock Efficiently scaling up crowdsourced video annotation.
\newblock {\em IJCV}, 2012.

\bibitem{youtube_faces}
L.~Wolf, T.~Hassner, and I.~Maoz.
\newblock Face recognition in unconstrained videos with matched background
  similarity.
\newblock In {\em {CVPR}}, 2011.

\bibitem{lbp_face2}
L.~Wolf, T.~Hassner, and Y.~Taigman.
\newblock Descriptor based methods in the wild.
\newblock In {\em Workshop on faces in `real-life' images: Detection,
  alignment, and recognition}, 2008.

\bibitem{SRC}
J.~Wright, A.~Y. Yang, A.~Ganesh, S.~S. Sastry, and Y.~Ma.
\newblock Robust face recognition via sparse representation.
\newblock {\em {PAMI}}, 2009.

\bibitem{face_recognition_context02}
R.~B. Yeh, A.~Paepcke, H.~Garcia-Molina, and M.~Naaman.
\newblock Leveraging context to resolve identity in photo albums.
\newblock In {\em JCDL}, 2005.

\bibitem{face_alignment03}
D.~Yi, Z.~Lei, and S.~Z. Li.
\newblock Towards pose robust face recognition.
\newblock In {\em CVPR}, 2013.

\bibitem{person_reid01}
D.~Yi, Z.~Lei, S.~Liao, and S.~Z. Li.
\newblock Deep metric learning for person re-identification.
\newblock In {\em ICPR}, 2014.

\bibitem{CASIA}
D.~Yi, Z.~Lei, S.~Liao, and S.~Z. Li.
\newblock Learning face representation from scratch.
\newblock {\em arXiv preprint arXiv:1411.7923}, 2014.

\bibitem{bird_pose_normalization}
N.~Zhang, R.~Farrell, and T.~Darrell.
\newblock Pose pooling kernels for sub-category recognition.
\newblock In {\em CVPR}, 2012.

\bibitem{panda}
N.~Zhang, M.~Paluri, M.~Ranzato, T.~Darrell, and L.~Bourdev.
\newblock Panda: Pose aligned networks for deep attribute modeling.
\newblock In {\em CVPR}, 2014.

\bibitem{person_recognition_poselets}
N.~Zhang, M.~Paluri, Y.~Taigman, R.~Fergus, and L.~Bourdev.
\newblock Beyond frontal faces: Improving person recognition using multiple
  cues.
\newblock In {\em {CVPR}}, 2015.

\bibitem{dksvd}
Q.~Zhang and B.~Li.
\newblock Discriminative k-svd for dictionary learning in face recognition.
\newblock In {\em CVPR}, 2010.

\bibitem{person_reid10}
R.~Zhao, W.~Ouyang, and X.~Wang.
\newblock Learning mid-level filters for person re-identification.
\newblock In {\em CVPR}, 2014.

\bibitem{Face_Recognition_Literature_Survey}
W.-Y. Zhao, R.~Chellappa, P.~J. Phillips, and A.~Rosenfeld.
\newblock Face recognition: A literature survey.
\newblock {\em ACM computing surveys}, 2003.

\bibitem{face_alignment06}
X.~Zhu, Z.~Lei, J.~Yan, D.~Yi, and S.~Z. Li.
\newblock High-fidelity pose and expression normalization for face recognition
  in the wild.
\newblock In {\em CVPR}, 2015.

\end{thebibliography}
}

\clearpage
\section*{Supplementary}
\setcounter{section}{0}
\def\thesection{\Roman{section}}
This supplementary document provides additional qualitative and quantitative results that provide further insights into the results discussed in the main paper. A more detailed description of the datasets is given in~\autoref{sec:supp_datasets} and the pose clusters discussed in~\autoref{sec:viewlearn} of the main paper are visualized in \autoref{sec:supp_pose_cluster_visualization}. Additional experimental results and visualizations that show the effectiveness of different components of our framework are given in \autoref{sec:supp_quantitative} and \autoref{sec:supp_qualitative}, respectively.

\section{Datasets}
\label{sec:supp_datasets}
\subsection{IMDB}
\begin{figure}[!t]
\centering
\includegraphics[height = 8cm,width = 8cm]{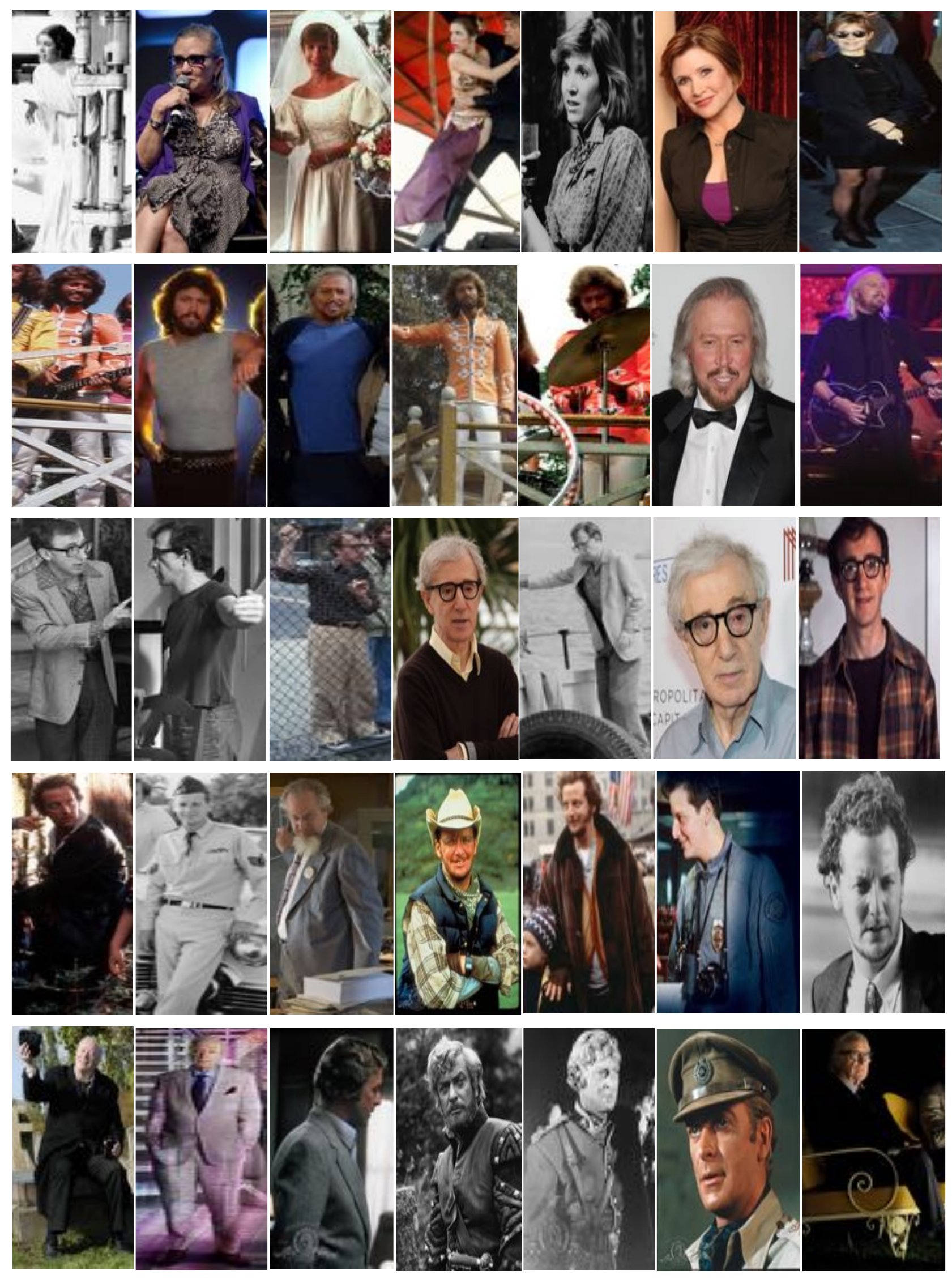}
\caption{{\bf IMDB:} Each row shows few images of an actor from the dataset. We used \textsc{IMDB} dataset to train classifiers for actor recognition in the Hannah movie.}
\label{fig:imdb_images}
\end{figure}

We created the \textsc{IMDB} database to train the actor classifier in the movie scenario. The scenario is different from most of the person recognition in that the test set contains a single movie with lesser variation in appearance between multiple instances of an actor in terms of age, style of clothing, etc. We assume that there are no labeled images within the movie and hence training data is not a part of the movie. To create a training set, the images are collected from the \textsc{IMDB} profile\footnote{\url{http://www.imdb.com/title/tt0091167/fullcredits}} of actors appearing in the movie, which are then manually cropped and annotated. Few images from \textsc{IMDB} database are shown in Figure~\ref{fig:imdb_images}. We relied on text tags associated with photos for annotation whenever the photos contain multiple confusing identities. Apart from illumination, resolution, and pose variations, there is a large age variations among \textsc{IMDB} instances. In addition, there is a large domain contrast between \textsc{IMDB} and Hannah test set in terms of lighting, camera and imaging conditions. This creates a more challenging setting to match identites between \textsc{IMDB} and Hannah instances.

\subsection{Soccer}
\begin{figure*}
\centering
\includegraphics[width = 0.88\linewidth]{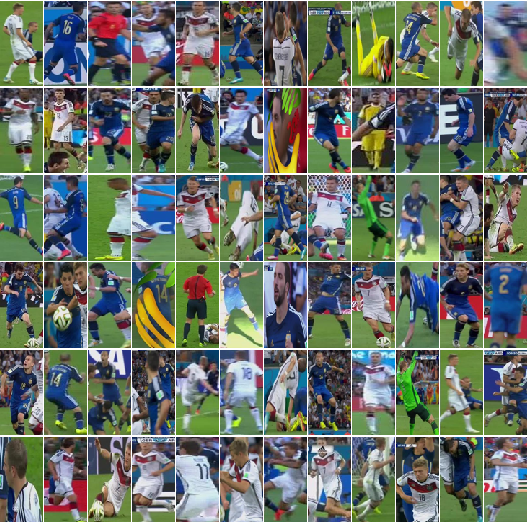}
\caption{Images from soccer dataset. It offers a challenging person recognition scenario due to low resolution, high occlusion, deformation and motion blur exhibited by soccer instances.}
\label{fig:soccer_dataset}
\end{figure*}

Soccer is another scenario where there are a significant number of frames in which the face is not visible and the subjects are often occluded by other players. We show more examples from our soccer dataset in Figure~\ref{fig:soccer_dataset}. In many instances, head is largely occluded, and in back-view unlike \textsc{PIPA} and Hannah instances, which contain visible head and torso regions. Also, soccer instances exhibit large body deformations, are of low resolution with significant blur. The soccer dataset therefore offers different kinds of challenges for recognition that are not seen in \textsc{PIPA} and Hannah.

\begin{figure}[!t]
\centering
\includegraphics[height = 10cm, width=8cm]{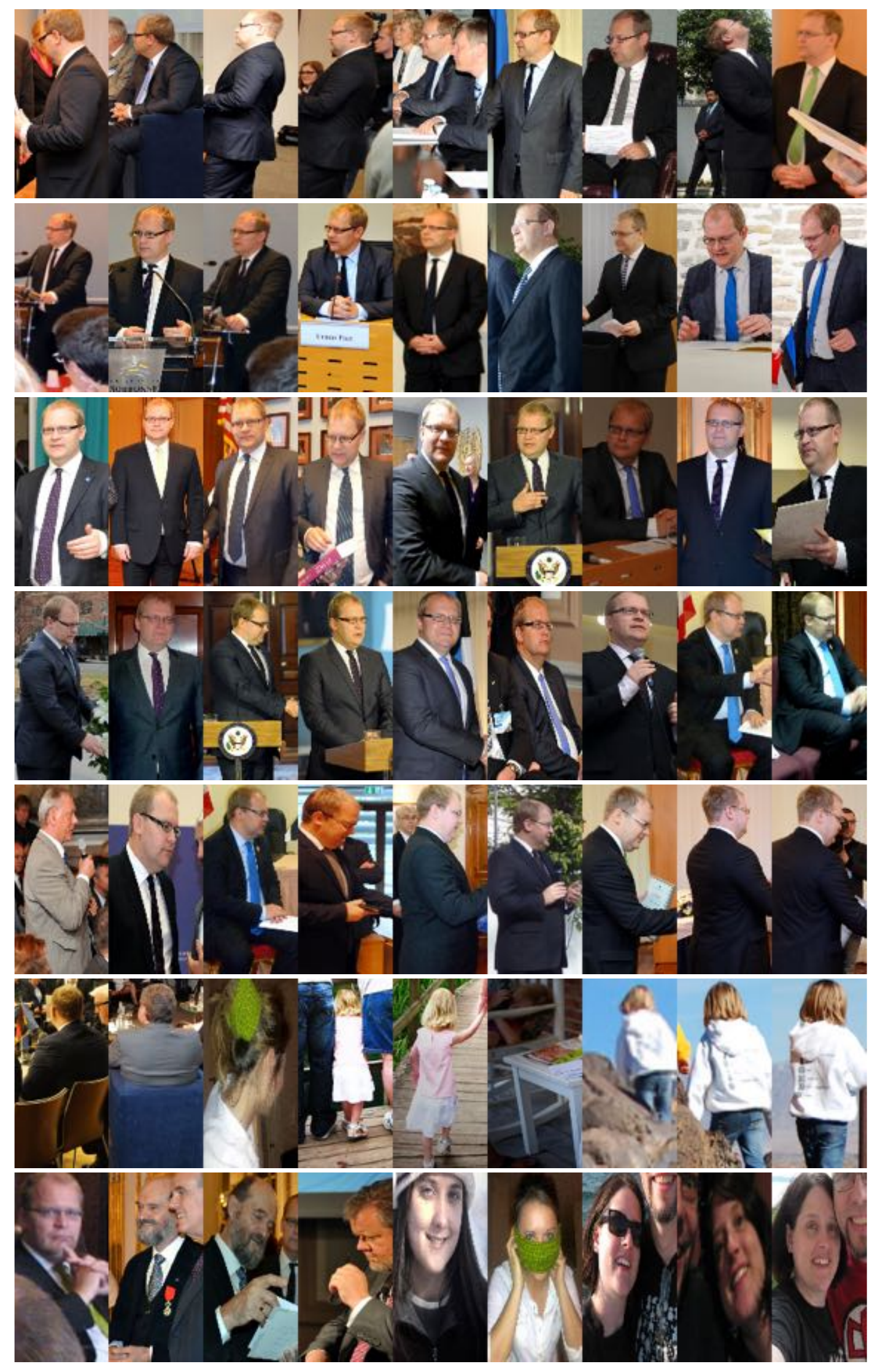}
\caption{{\bf Pose clusters:} Each row from top to bottom shows people from \textsc{PIPA} with particular body orientation clustered using orientation and keypoint visibility features. }
\label{fig:pose_clusters}
\end{figure}

\section{Pose clusters}
\label{sec:supp_pose_cluster_visualization}
We obtain a set of prominent views to facilitate pose-specific representations as discussed in \autoref{sec:viewlearn}. To achieve this, we annotated $14$ body keypoints for $29{,}223$ \textsc{PIPA} train instances which are then used for clustering. More examples of our pose clusters are shown in Figure~\ref{fig:pose_clusters}. Each row from top to bottom contain images from {\tt right}, {\tt semi-right}, {\tt frontal}, {\tt semi-left}, {\tt left}, {\tt back} and {\tt partial body} views. The orientation and keypoint visibility features produced tight clusters containing images with particular body orientation. The last cluster captures the instances with partial upper body such as head or shoulder, etc, in the images that are commonly seen in social media photos and movies. While we considered seven prominent views in this work, we note that generating a large number of views can be helpful, provided there are enough training samples in each cluster to train the convnets.

\section{Quantitative Results and Analysis}
\label{sec:supp_quantitative}
\begin{figure}
\centering
\includegraphics[width = 0.9\linewidth]{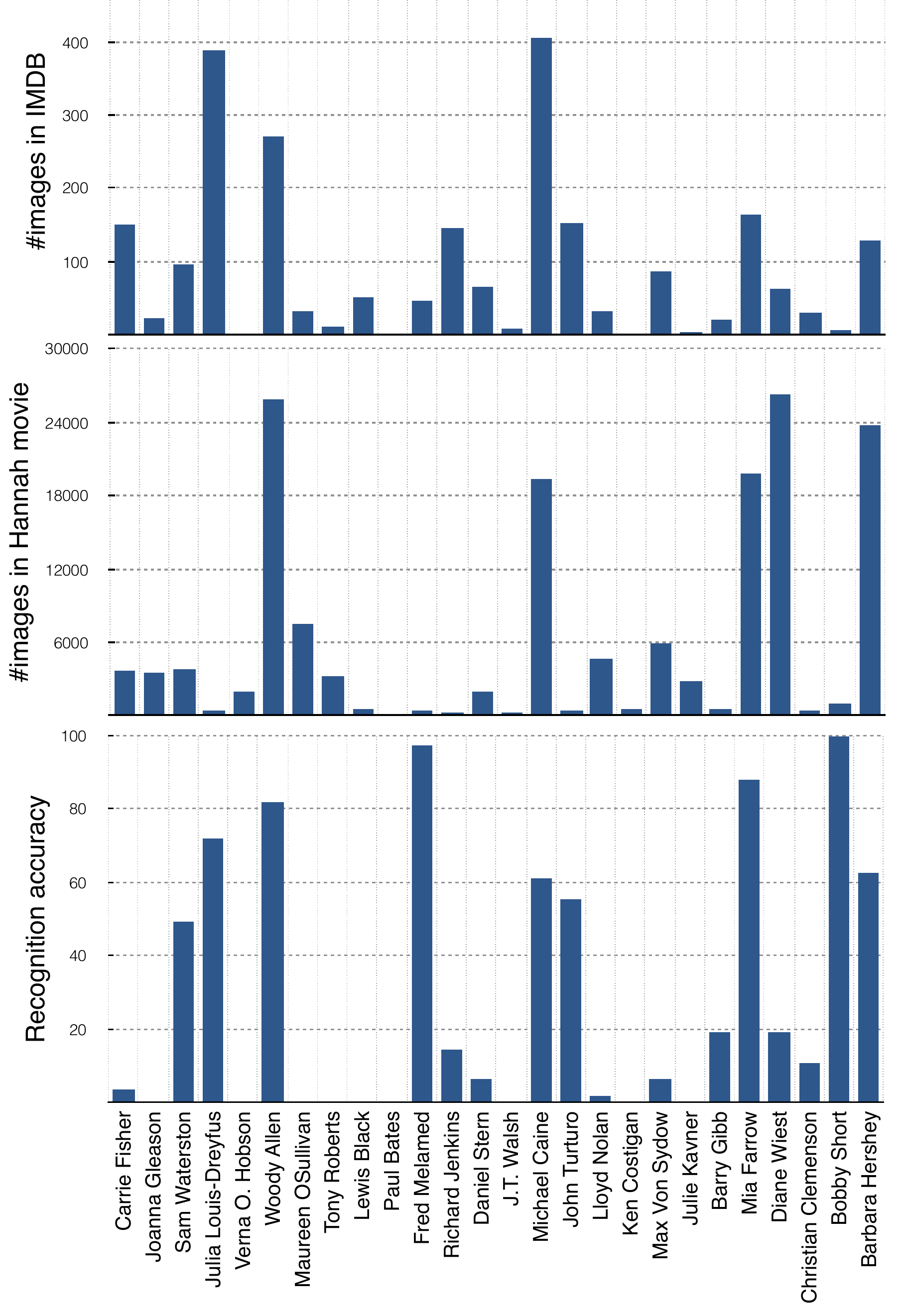}
\caption{Number of images for each actor in (top) \textsc{IMDB} and (middle) Hannah movie test set. We show the (bottom) recognition performance of each actor on the test set.}
\label{fig:imdb_hannah_stat}
\end{figure}

\begin{figure}
\centering
\includegraphics[width = 0.9\linewidth]{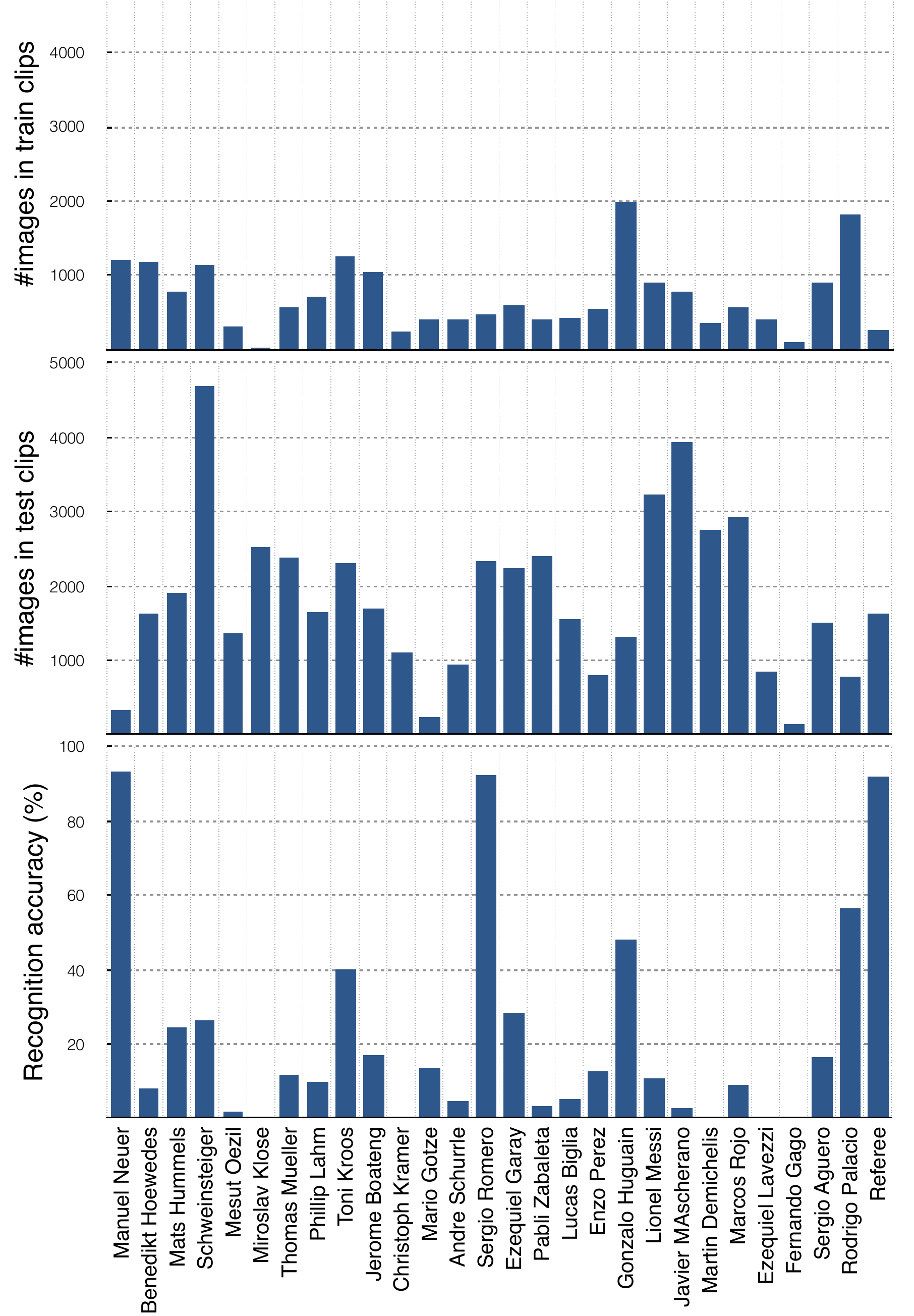}
\caption{Number of images for each player in the training (top) and test (middle) split of Soccer dataset. We also show the (bottom) recognition performance of each player.}
\label{fig:soccer_stat}
\end{figure}

We provide more insightful results that help to understand merits and challenges of different recognition settings that are considered.


{\bf Recognition per subject:} 
Figure~\ref{fig:imdb_hannah_stat} shows the number of images for each actor in \textsc{IMDB} and Hannah test sets along with their individual recognition performances. We observe that, for those subjects with sufficiently large number of training instances ({\em Michael Caine, Barbara Harshey, Woody Allen, Julia Louis-Dreyfus, and Mia Farrow}), the performance is high as expected. For subjects with less than $20$ training instances, the performance is very low. However, whenever there is a large difference in age between train and test instances ({\em Carrie Fisher, Dianne West, Richard Jenkins}), the performance is poor despite having enough training examples.

Similarly, we show the statistics of soccer players along with their individual performances in Figure~\ref{fig:soccer_stat}. We see a similar trend of high performance for subjects ({\em Gonzalo Huguain} and {\em Rodrigo Palacio}) with sufficient training instances. We also observe a near $100$\% accuracy for goal keepers ({\em Manuel Neuer} and {\em Sergio Romero}) and the referee due to clothing cues, which are discussed next.

\begin{figure}
\centering
\includegraphics[width = 0.95\linewidth]{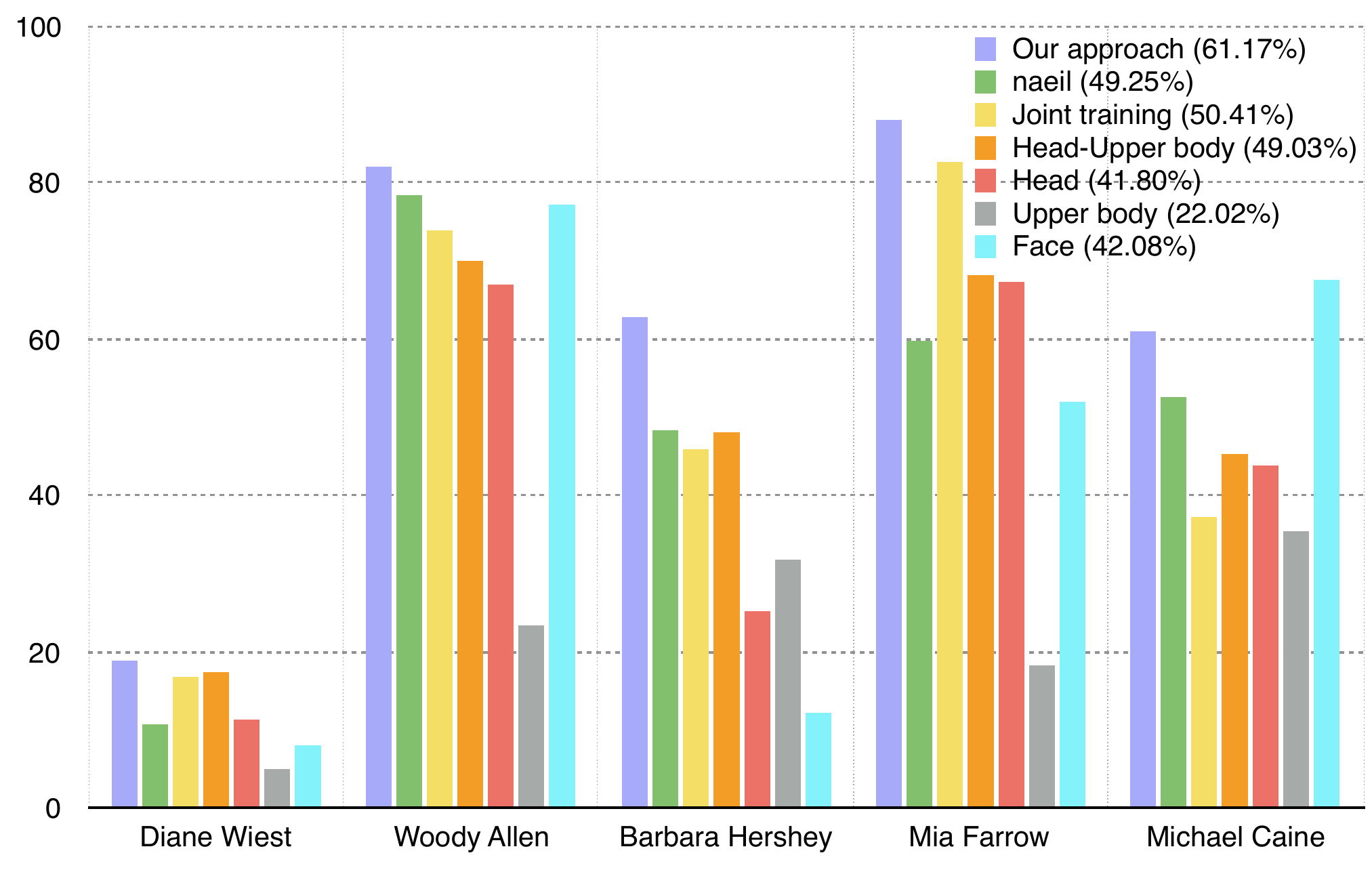}
\caption{Recognition performance of five lead actors in Hannah dataset.}
\label{fig:hannah_top5}
\end{figure}

\begin{figure}
\centering
\includegraphics[width = 0.95\linewidth]{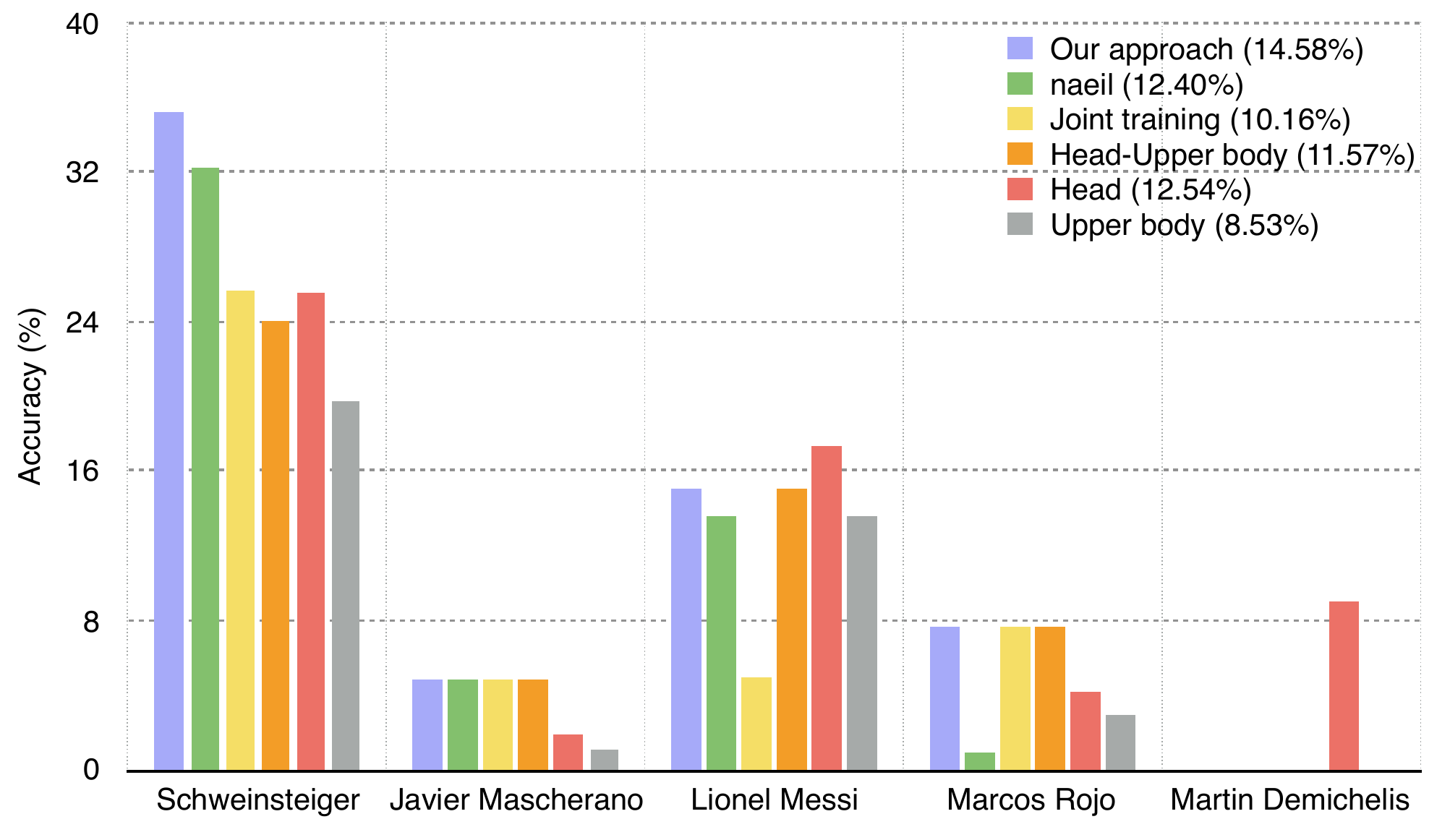}
\caption{Recognition performance of five most occurring players in Soccer dataset.}
\label{fig:soccer_top5}
\end{figure}

{\bf Recognition performance of top subjects:} 
We compare the recognition performance of various approaches on $5$ most occurring movie and soccer subjects in Figure~\ref{fig:hannah_top5} and Figure~\ref{fig:soccer_top5}, respectively. Our approach reaches an accuracy of $61.17$\% on top actors, which is significantly better than {\tt naeil}. Note that the overall performance of {\tt naeil} with $17$ models is comparable to head and upper body. Unlike photo-albums, clues such as scene and human attributes like age, glasses, and hair color are less useful in the movie setting. For actors with less change in appearance over time ({\em Michael Caine} and {\em Woody Allen}: See row three and five in Figure~\ref{fig:imdb_images}), face is found to be extremely informative and robust compared to head.

On the soccer dataset, the overall performance is poor for all the approaches. This suggest to develop better representations that are able to recognize people at a distance.

\begin{figure}
\centering
\includegraphics[scale=0.46]{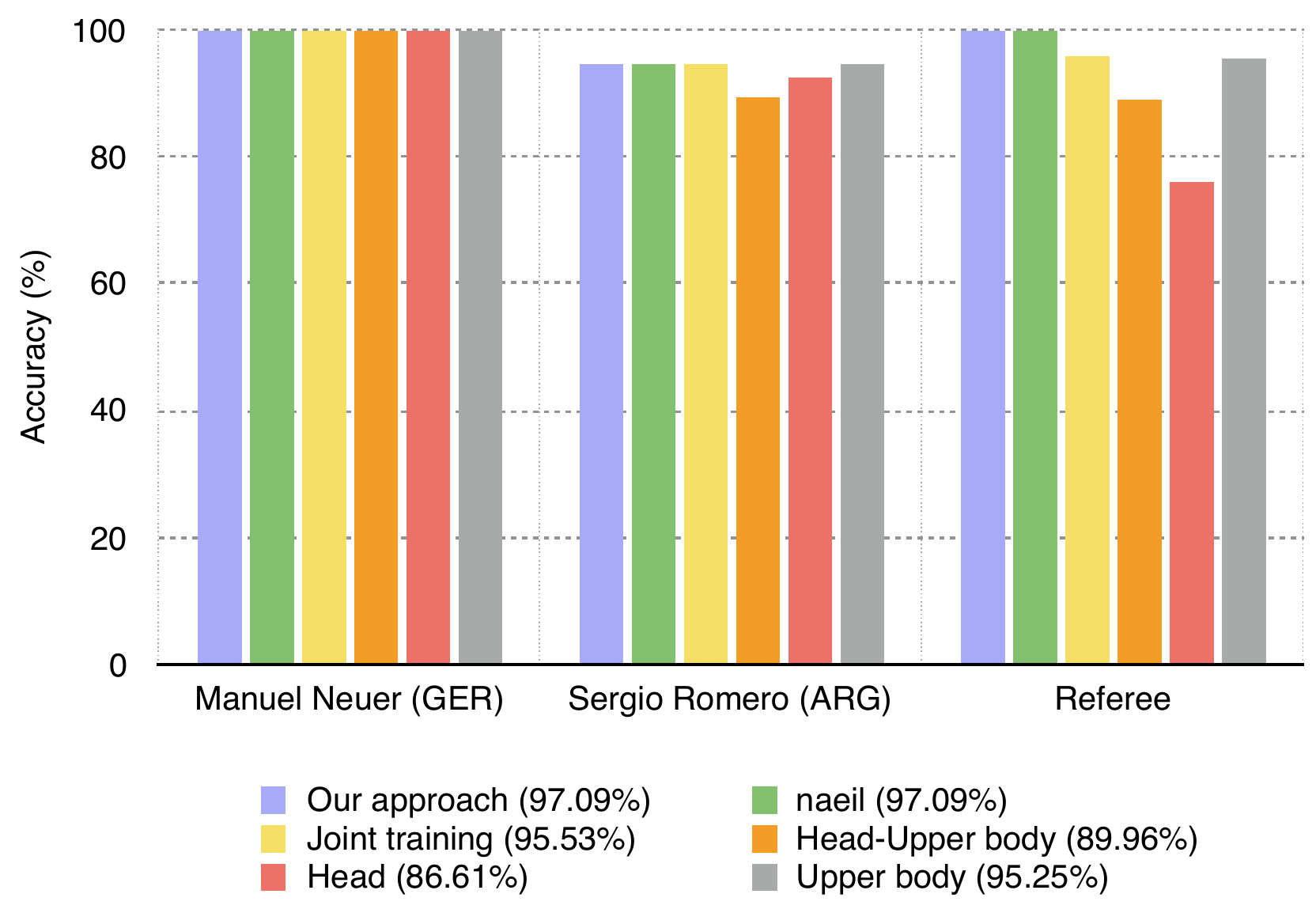}
\caption{Effect of clothing on recognition.}
\label{fig:soccer_clothing}
\end{figure}

{\bf How informative is clothing?} Though it is intuitively obvious that clothing helps in recognition, a qualitative evaluation is not done previously. We perform such a study using the soccer dataset. We show the performance of different approaches on three subjects ({\em Manuel Neuer}, {\em Sergio Romero} and {\em Referee}) with unique clothing in Figure~\ref{fig:soccer_clothing}. The first two subjects are the goal keepers of the Germany and Argentina, respectively.

As seen in Figure~\ref{fig:soccer_clothing}, upper body region, which is often less informative compared to head, outperforms head by a large margin due to clothing. The concatenation of head and upper body obtained through separate training is worse than upper body feature alone. On the other hand, the concatenation of features using jointly trained model is more robust and performs much better as it provide more flexibility to focus on selective regions. Finally, the overall performance of pose aware models and {\tt naeil} are identical.

It is interesting to note that, convnets that are trained for identity recognition can distinguish clothing without any explicit modeling or hand-crafted features \cite{soccer02}.

\begin{figure*}
\begin{center}
\subfigure{
 \includegraphics[scale=0.6]{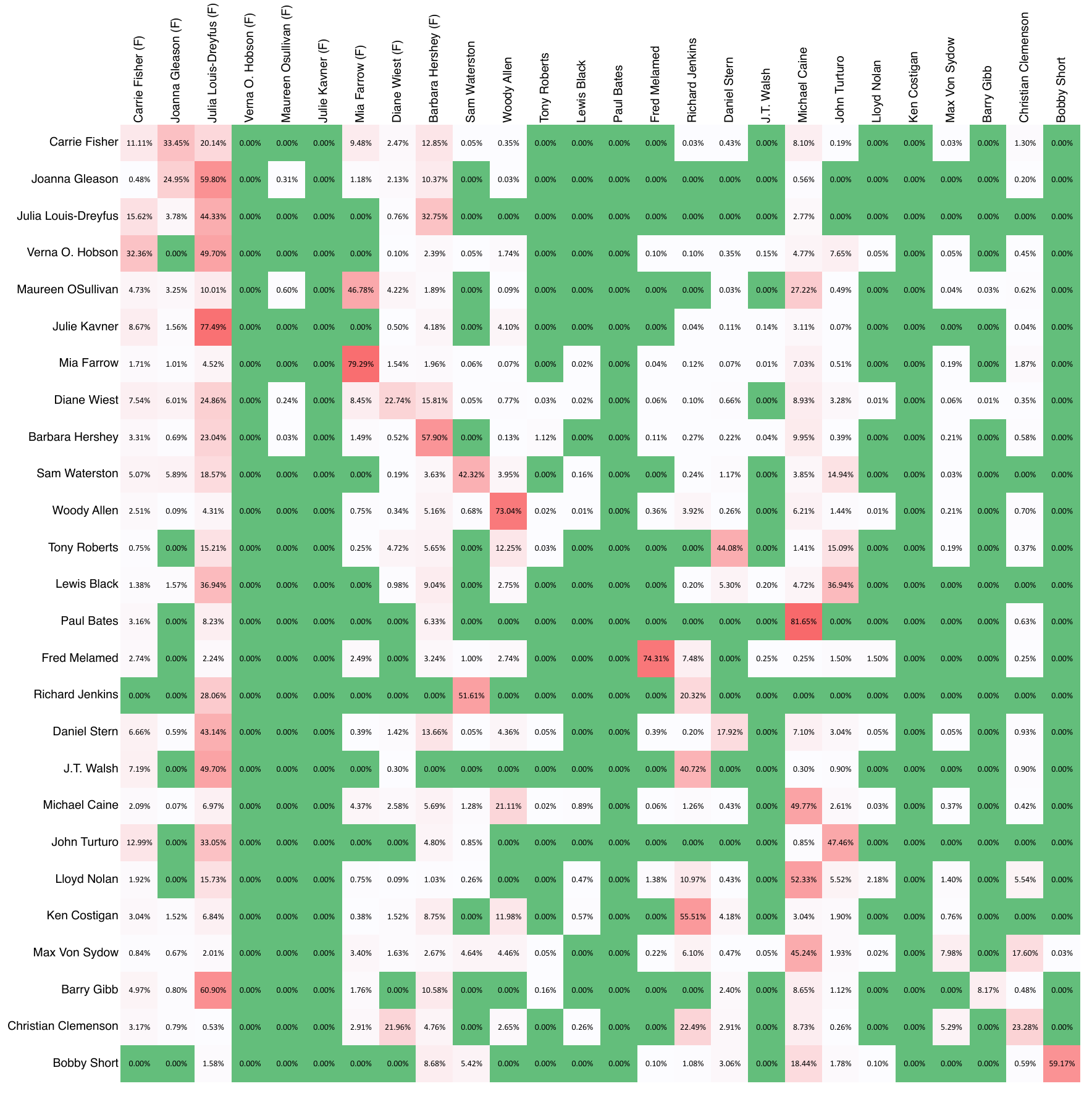}
}

\subfigure{
\includegraphics[scale=0.6]{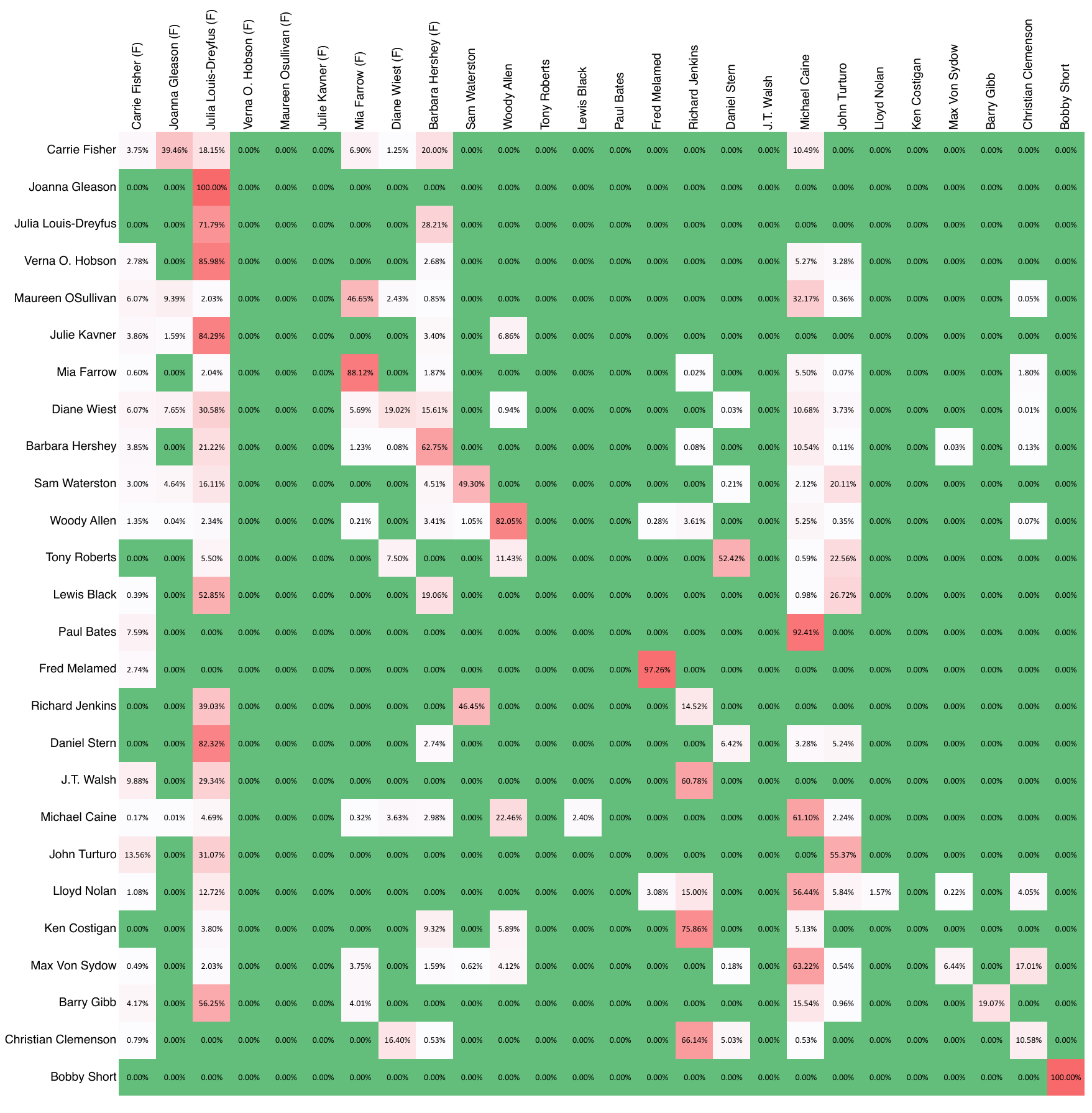}
}
\end{center}
\vspace*{-0.2in}
\caption{{\bf Confusion matrix} on Hannah dataset (top) with and (bottom) without tracking.}
\label{fig:hannah_confusion}
\end{figure*}

\begin{figure*}
\begin{center}
\subfigure{
\includegraphics[scale=0.6]{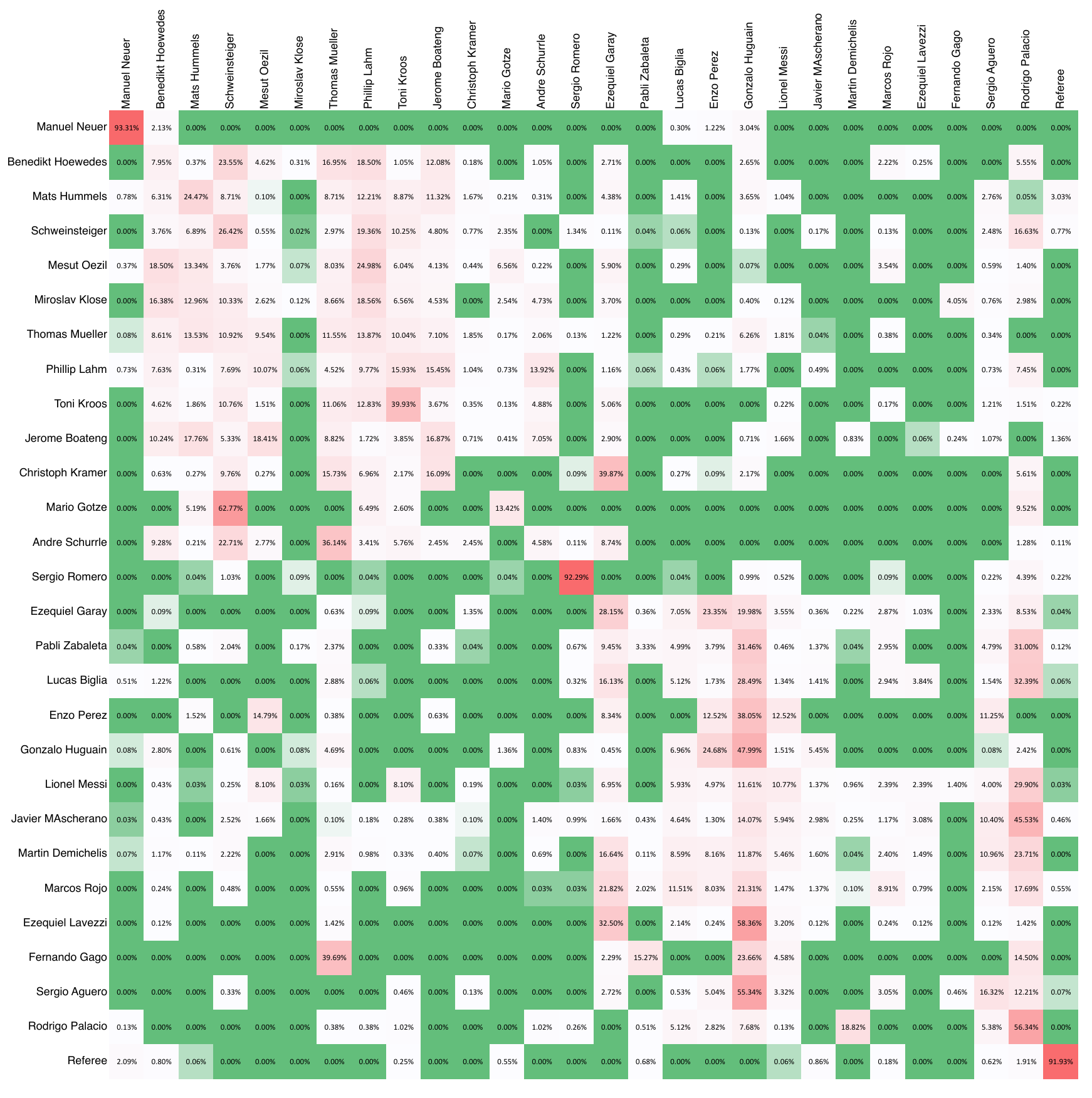}
}
\subfigure{
\includegraphics[scale=0.6]{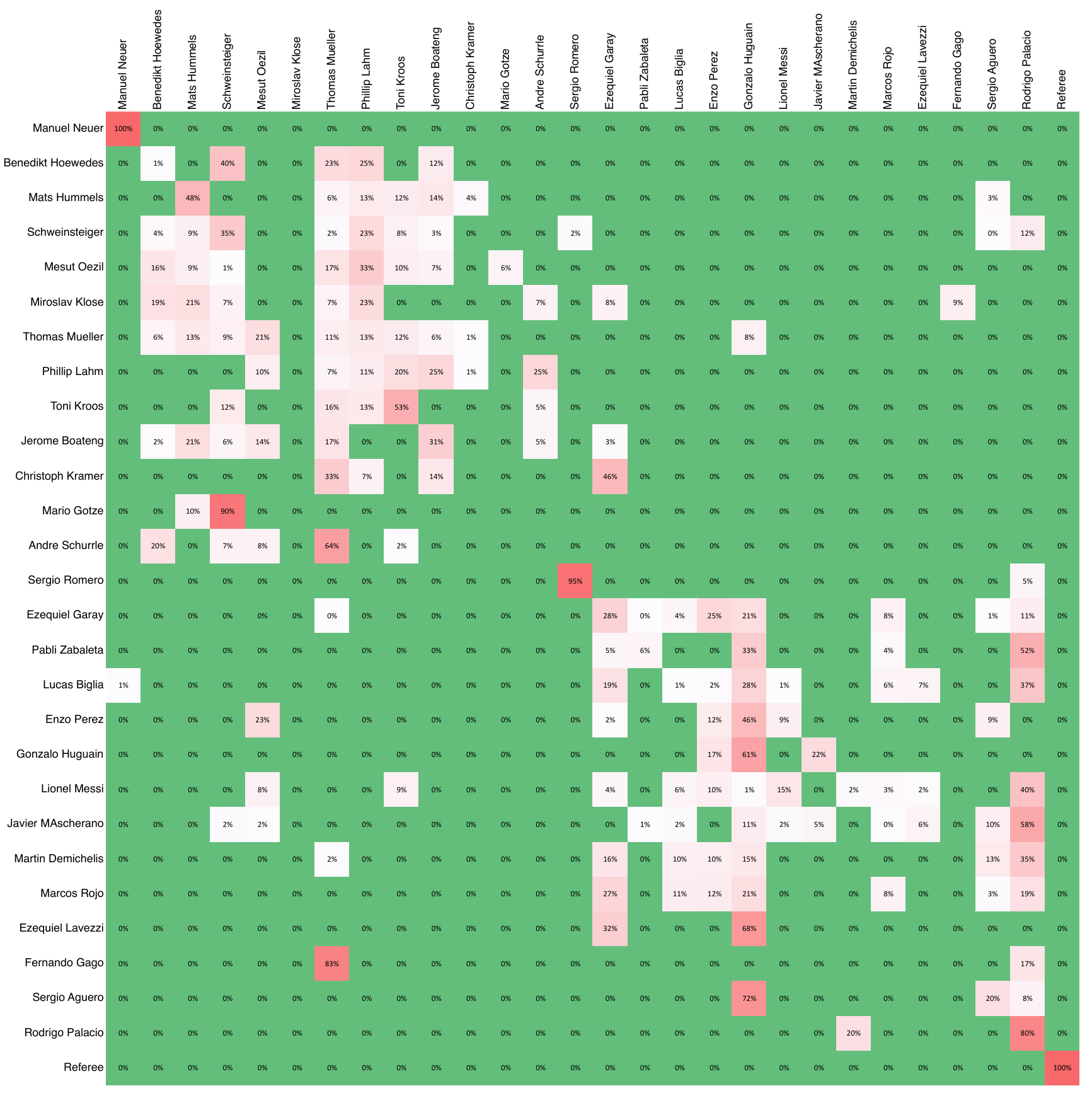}
}
\end{center}
\vspace*{-0.2in}
\caption{{\bf Confusion matrix} on Soccer dataset (top) with and (bottom) without tracking.}
\label{fig:soccer_confusion}
\end{figure*}

{\bf Confusion between identities:} We show the recognition confusion matrix for Hannah and Soccer datasets in Figure~\ref{fig:hannah_confusion} and Figure~\ref{fig:soccer_confusion} respectively, with and without tracking. We notice two important points related to gender and clothing. As seen from Figure~\ref{fig:hannah_confusion}, female subjects are mostly getting confused with female subjects, and similarly the male subjects are confused with male subjects. In Figure~\ref{fig:soccer_confusion}, we notice that players from each team are mislabeled with the members from the same team. These studies show the effectiveness of convnets in capturing human attributes without any explicit training.
Finally, majority voting over a track helps to produce consistent predictions.

\begin{figure}
\centering
\includegraphics[width = 0.8\linewidth]{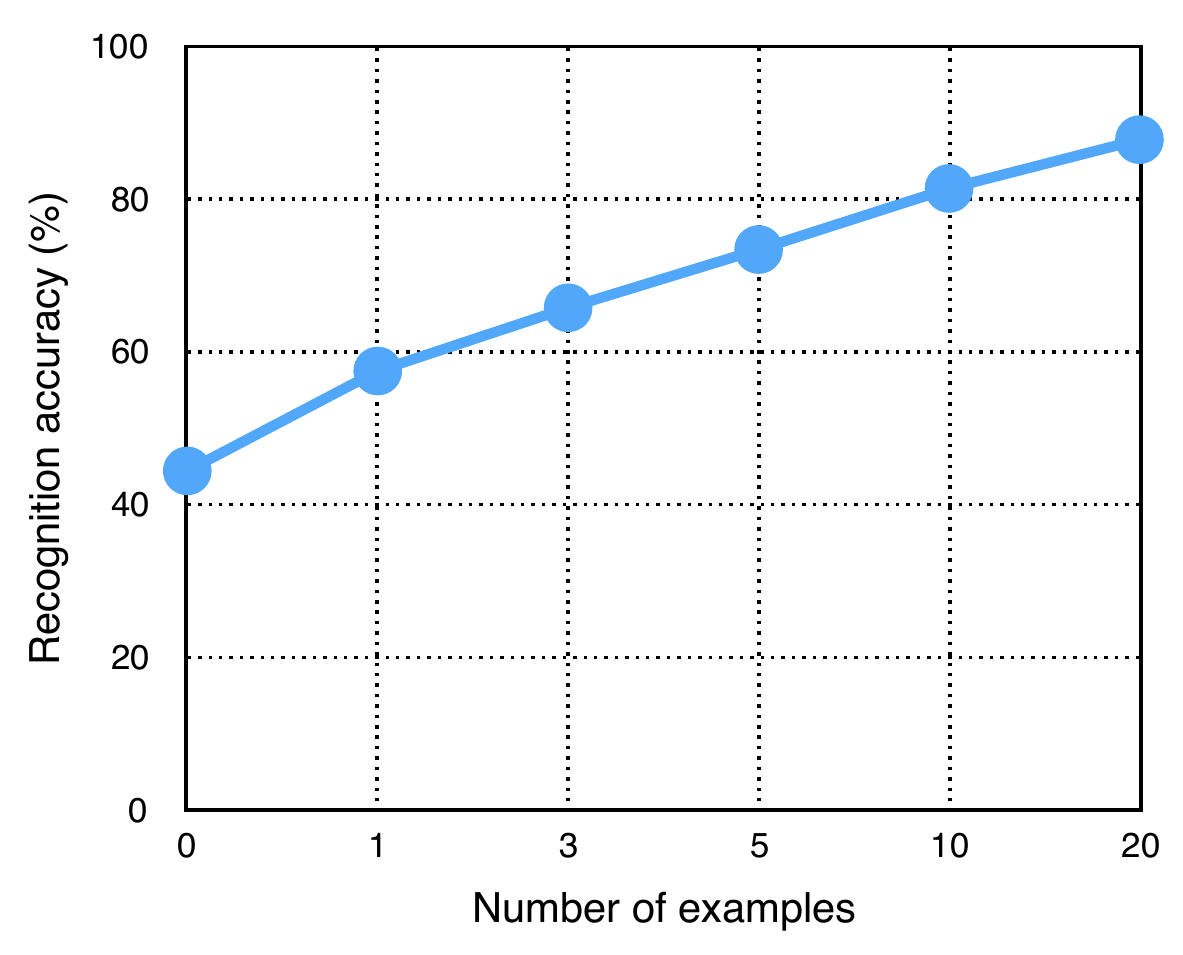}
\caption{Recognition performance of Hannah movie set using \textsc{IMDB} plus samples from the Hannah test set.}
\label{fig:hannah_acc_vs_num_examples}
\end{figure}

{\bf Domain gap:} To understand the effect of domain contrast between train and test instances, we conduct an experiment adding different number of Hannah instances per subject to the \textsc{IMDB} training gallery. The results are shown in Figure~\ref{fig:hannah_acc_vs_num_examples}. As seen from the graph, the addition of even a few instances from the test domain results in a very large improvement in the recognition performance.

\section{Qualitative Results}
\label{sec:supp_qualitative}
\begin{figure*}
\centering
\includegraphics[scale=0.55]{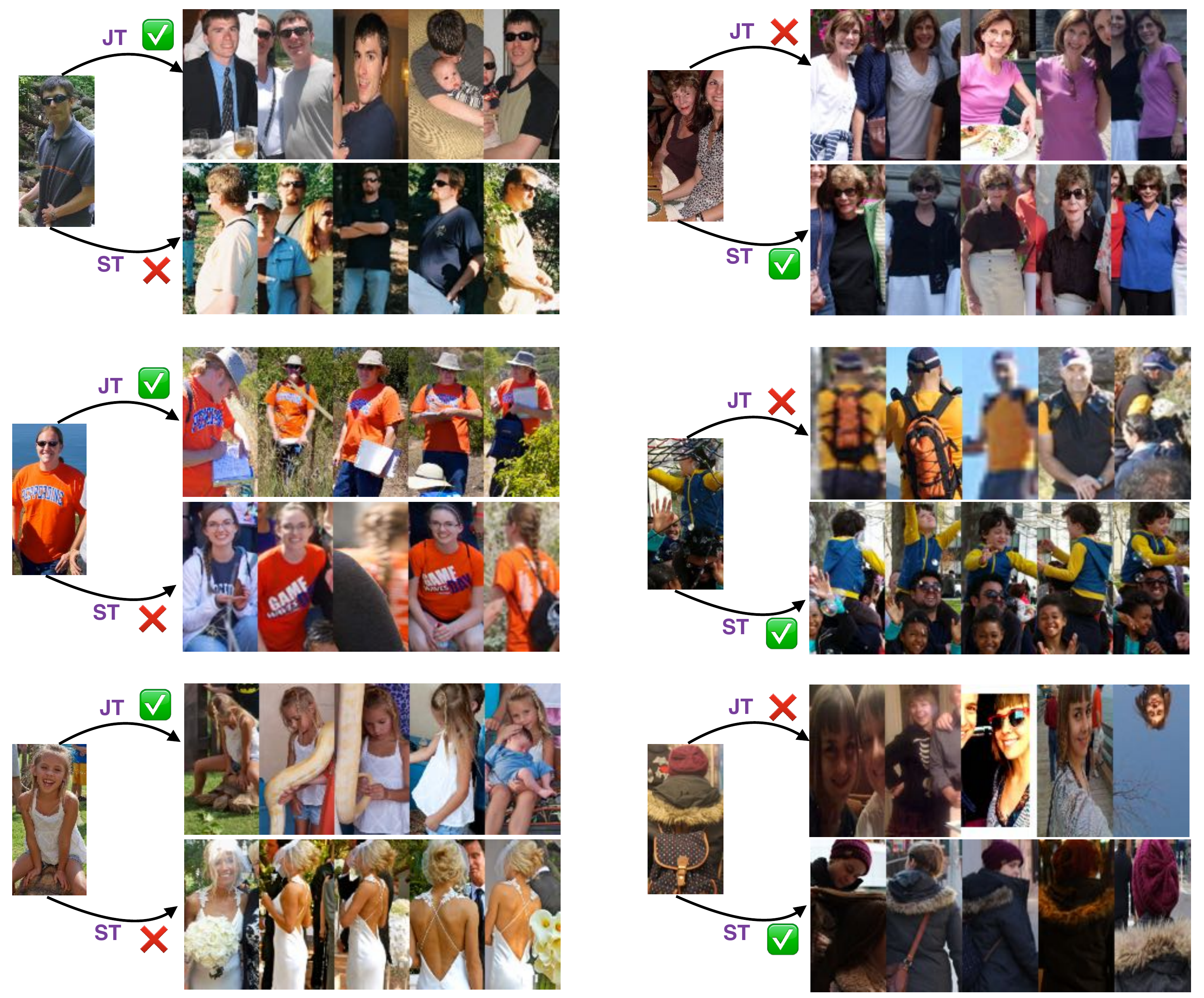}
\caption{Success and failure cases of separate and joint training of body regions on \textsc{PIPA} dataset. Column one shows the test images and the column two shows the training images belonging to the predicted subject.
(Left) shows the success and failure case of joint training (JT) and separate training (ST), respectively and the reverse is shown in (right).}
\label{fig:pipa_success_failure_separate_joint}
\end{figure*}

\begin{figure*}
\begin{center}
\subfigure{
 \includegraphics[scale=0.5]{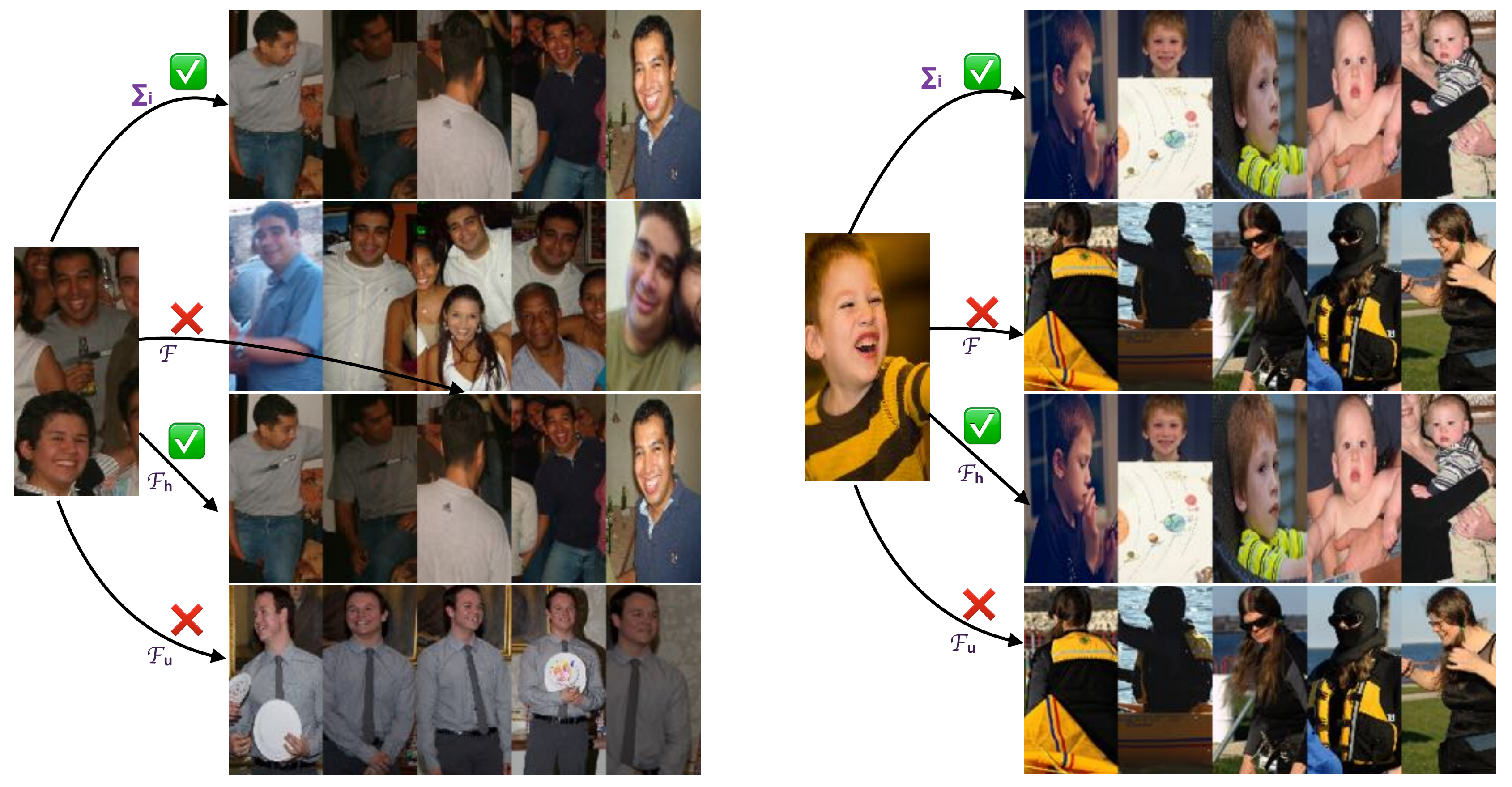}
}
\subfigure{
 \includegraphics[scale=0.5]{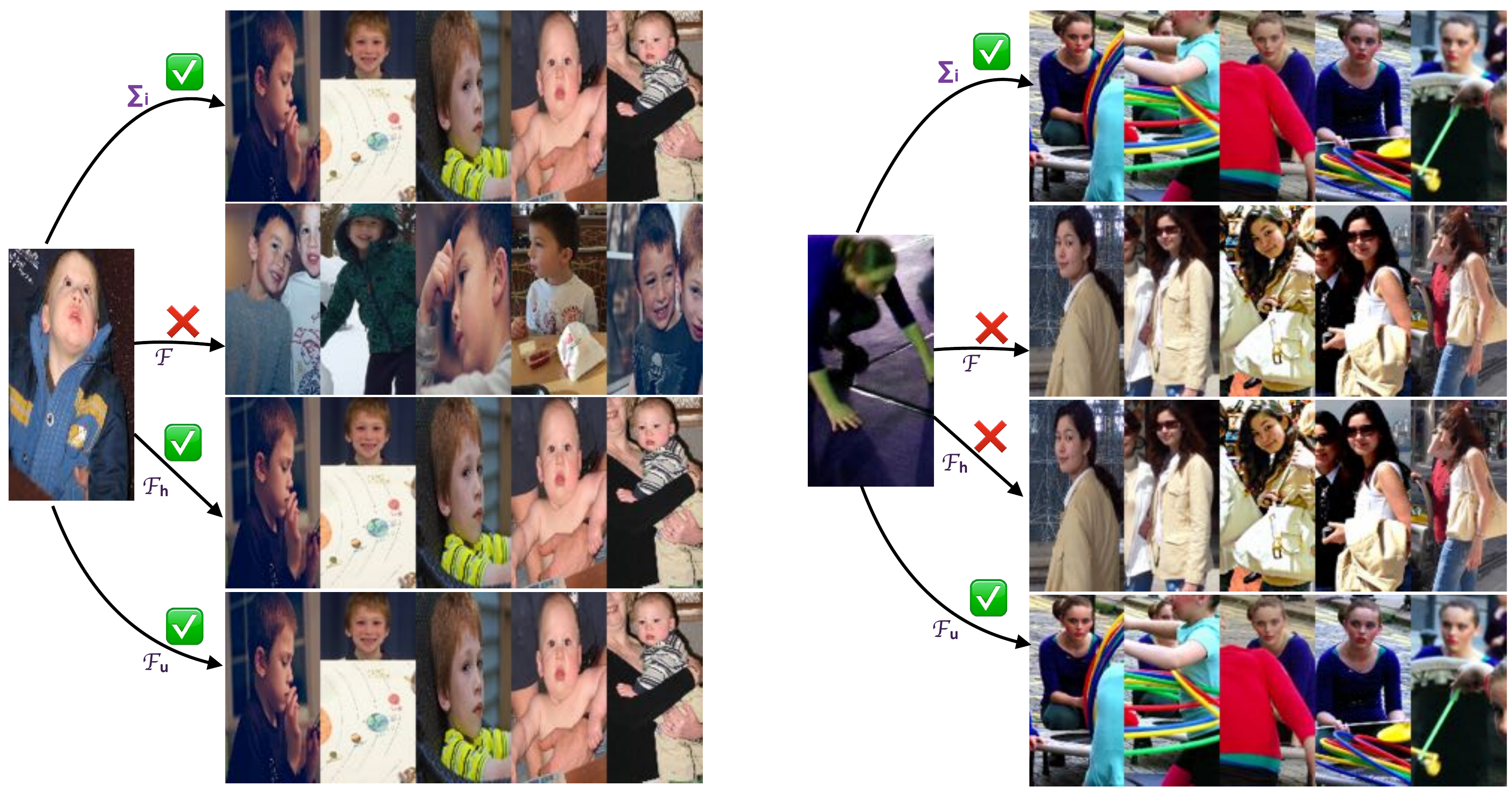}
}
\end{center}
\caption{{\bf Effectiveness of multiple classifiers from each {\it PSM}:} Column one shows the \textsc{PIPA} test images and the column two shows the training images belonging to the predicted subject using different approaches. The four approaches considered are the classifiers trained on head ($\mathcal{F}_h$) and upper body ($\mathcal{F}_u$) features, a classifier trained on concatenated head and upper body ($\mathcal{F}$) feature, and linear combination of three classifiers ($\sum_i$) trained on these features. It clearly shows that it is advantageous to consider individual classifiers trained on regional features and their combination for improved performance.}
\label{fig:pipa_success_failure_linear_comb}
\end{figure*}

\begin{figure*}
\includegraphics[height=20cm,width=17cm]{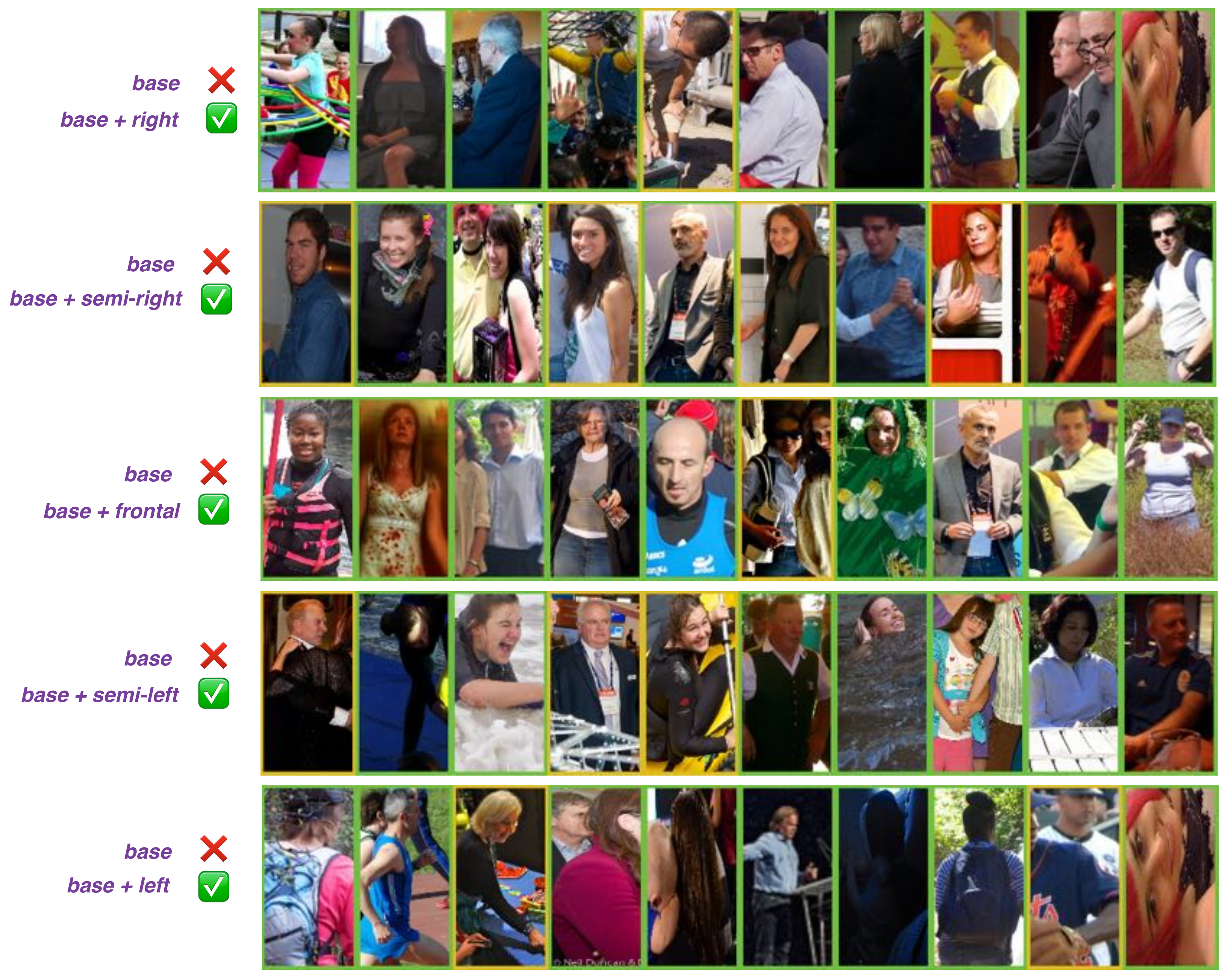}
\caption{{\bf Success cases of pose-specific models ({\it PSM}s) on \textsc{PIPA} dataset.}  Each row shows the success predictions of our approach where the improvement is obtained primarily due to the specific-pose model \ie, {\tt base} model wrongly predicts but {\tt base} + correct {\tt PSM} predicts correctly. Green and yellow boxes indicate the success and failure result of {\tt naeil} respectively.}
\label{fig:pipa_pose_success}
\end{figure*}

\begin{figure*}
\includegraphics[height=20cm,width=17cm]{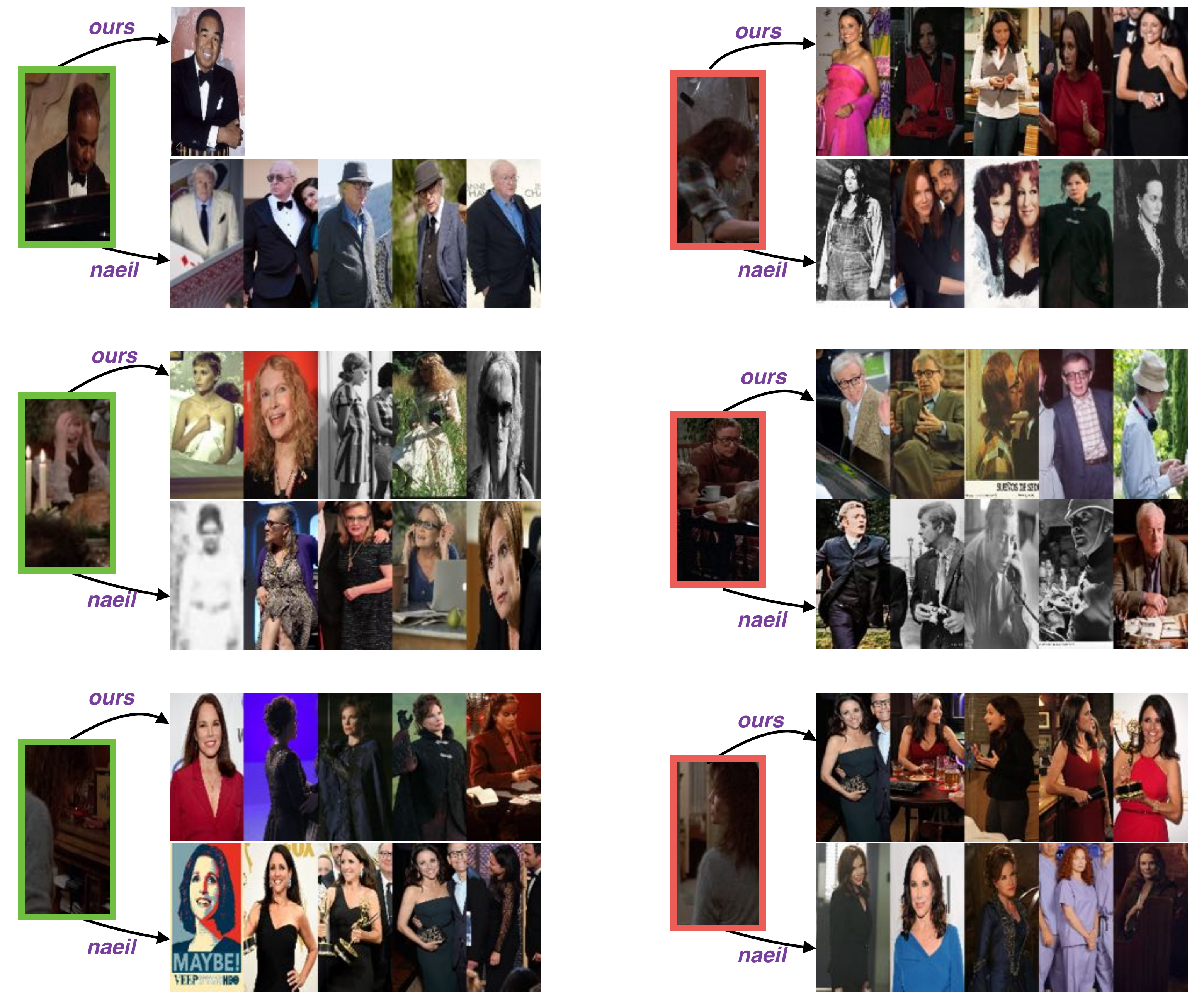}
\caption{Comparison of our approach with {\tt naeil} on Hannah dataset. Column one shows the test images and the column two shows the training images belonging to the predicted subject.
(Left) in green shows the success case of our approach and the failure case of {\tt naeil}. (Right) in red shows the failure case of our approach and the success case of {\tt naeil}.}
\label{fig:hannah_success_failure}
\end{figure*}

\begin{figure*}
\includegraphics[height=20cm,width=17cm]{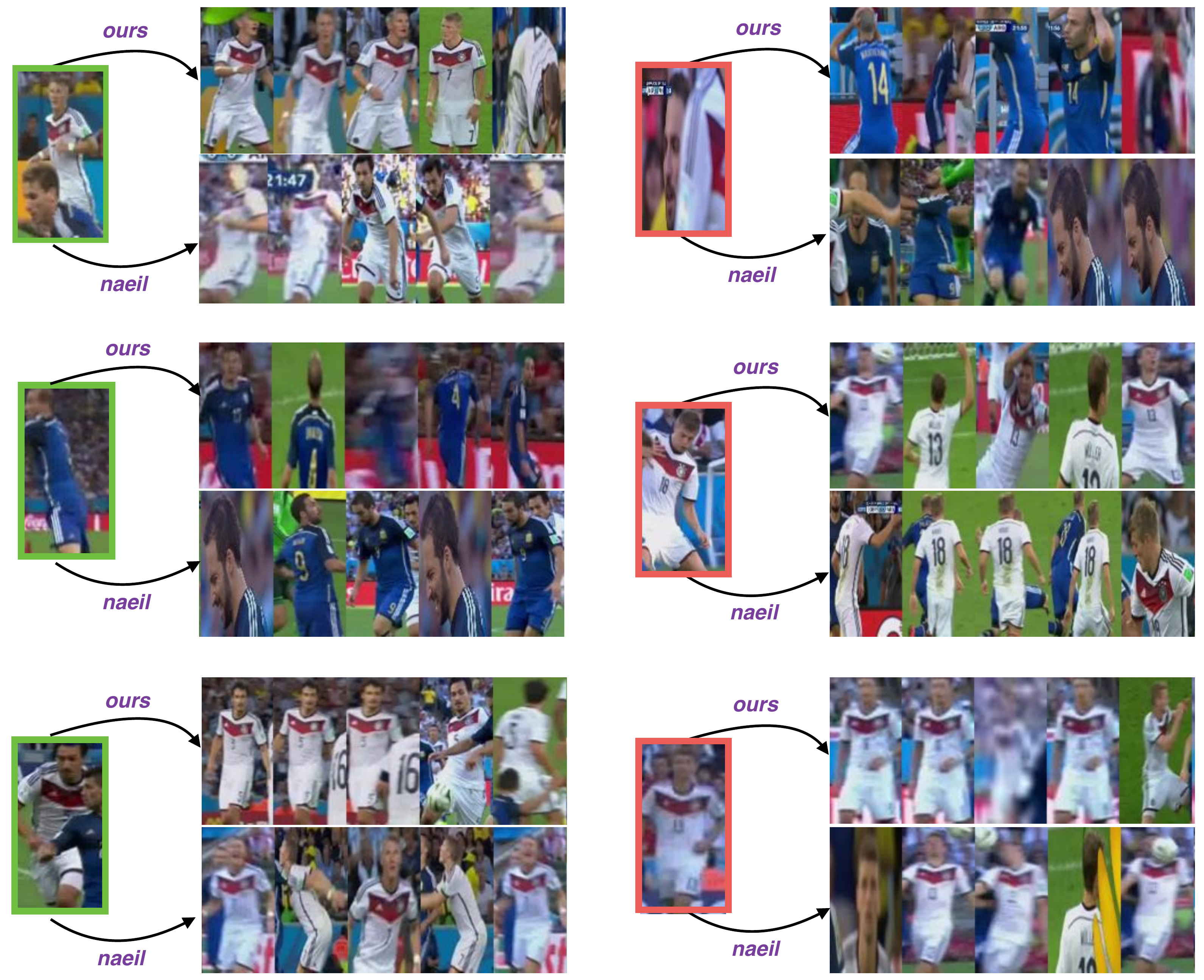}
\caption{Comparison of our approach with {\tt naeil} on Soccer dataset. Column one shows the test images and the column two shows the training images belonging to the predicted subject.
(Left) in green shows the success case of our approach and the failure case of {\tt naeil}. (Right) in red shows the failure case of our approach and the success case of {\tt naeil}.}
\label{fig:soccer_success_failure}
\end{figure*}

We show some qualitative results in Figures~\ref{fig:pipa_success_failure_separate_joint} to \ref{fig:soccer_success_failure}. Figure~\ref{fig:pipa_success_failure_separate_joint} shows the success and failure cases of joint training and separate training of body regions. We notice an over-influence of clothing while using separately trained and concatenated regional features, compared to the jointly training features. In Figure~\ref{fig:pipa_success_failure_linear_comb}, we show the effectiveness of using multiple classifiers from each {\it PSM}. As seen in the figure, the concatenated head and upper body features ($\mathcal{F}$) may predict incorrect labels even when one (or two) of these features predict correctly, due to the over influence of less informative body region. Combining these three features is found to be more robust.

We show the top scoring predictions obtained from each pose-specific {\it PSM} in Figure~\ref{fig:pipa_pose_success}. It clearly shows how each {\it PSM} helps in the prediction of instances in that particular pose when the base model is unable to predict correctly. Finally, we show the success and failure cases of our approach on Hannah and Soccer datasets in Figure~\ref{fig:hannah_success_failure} and Figure~\ref{fig:soccer_success_failure} respectively, and compare with the {\tt naeil}.

\end{document}